\documentclass[10pt]{amsart}
\usepackage[usenames,dvipsnames]{xcolor}
\usepackage{lmodern}
\usepackage[T1]{fontenc}
\usepackage{amsmath,amsthm,amssymb,color,enumerate,fancyhdr,graphicx,multirow,stmaryrd,verbatim}
\usepackage[round]{natbib}
\usepackage[multiple]{footmisc}

\usepackage{hyperref}\hypersetup{colorlinks=true,linkcolor=MidnightBlue,citecolor=MidnightBlue,urlcolor=MidnightBlue}

\usepackage{caption}

\usepackage{tikz}
\usetikzlibrary{arrows,backgrounds,calc,fit,positioning,shapes}

\theoremstyle{definition}

\newcommand{\commentout}[1]{}

\newcommand*{\QEDP}{\hfill\ensuremath{\Box}}

\DeclareMathOperator{\Expec}{\mathbf{E}}

\DeclareRobustCommand{\VAN}[3]{#2} % proper Dutch 'van/de' capitalisation

\begin{document}

\title[The no-free-lunch theorems]{The No-Free-Lunch Theorems of \\ Supervised Learning}
\author[Sterkenburg and Gr\"unwald]{Tom F.\ Sterkenburg and Peter D.\ Gr\"unwald}
\date{June 3, 2021}
\address{Munich Center for Mathematical Philosophy, LMU Munich}
\email{tom.sterkenburg@lmu.de}
\address{Machine Learning Group, Centrum Wiskunde \& Informatica, Amsterdam \newline \hspace*{12pt}Mathematical Institute, Leiden University}
\email{pdg@cwi.nl}
\thanks{This is the final version, that appeared in \emph{Synthese} 199:9979--10015 (2021), \href{https://doi.org/10.1007/s11229-021-03233-1}{doi: 10.1007/s11229-021-03233-1}. For helpful comments we would like to thank Gordon Belot, Kathleen Creel, Sam Fletcher, Bas van Fraassen,  Timo Freiesleben, Konstantin Genin,  Daniel Herrmann,  Wouter Koolen, J\"urgen Landes, Jonathan Livengood, Daniel Malinsky, Conor Mayo-Wilson, Eric-Jan Wagenmakers, and the anonymous reviewers. Tom Sterkenburg was supported by the Deutsche Forschungsgemeinschaft (\textsc{DFG}, German Research Foundation)---Projektnummern 437206810, \emph{Die Epistemologie der Statistischen Lerntheorie}; 432308570, \emph{Grundlagen, Anwendungen \& Theorie der Induktiven Logik}. Peter Gr\"unwald was supported by the Dutch Research Council (NWO) via research programme 617.001.651, \emph{Safe Bayesian Learning}. }

\begin{abstract}
The no-free-lunch theorems promote a skeptical conclusion that all possible machine learning algorithms equally lack justification. But how could this leave room for a learning theory, that shows that some algorithms are better than others? Drawing parallels to the philosophy of induction, we point out that the no-free-lunch results presuppose a conception of learning algorithms as purely data-driven. On this conception, every algorithm must have an inherent inductive bias, that wants justification. We argue that many standard learning algorithms should rather be understood as model-dependent: in each application they also require for input a model, representing a bias. Generic algorithms themselves, they can be given a model-relative justification.
\end{abstract}

\maketitle

%TOM to PETER: ik heb de introductiE grotendeels herschreven, ook in reactie op jouw commentaar. 
\section{Introduction}
The  no-free-lunch (\textsc{NFL}) theorems of supervised learning \citep{Wol92cs,Wol96nc,Sch94icml} are an influential
%PETER: celebrated ? (famous -> only for people? 
%TOM to PETER: te positief? compromis: "influential" )
collection of impossibility results in machine learning. Computer scientists have ranked these results ``among the most important theorems in statistical learning'' \citep[695]{LuxSch11incl}, while some philosophers have read them as ``a radicalized version of Hume's induction skepticism'' \citep[825, 830]{Sch17pos}.

In a nutshell, the results say---or rather, are usually \emph{interpreted} as saying---that we cannot formally justify our machine learning algorithms. That is, we cannot formally ground our conviction that some learning algorithms are more sensible than others: that we have reason to think some algorithms perform better in attaining the epistemic goals that we designed them to attain. In Wolperts original interpretation, ``all learning algorithms are equivalent'' (\citealp[129]{Wol95inc}; \citeyear[35]{Wol02inc}), so that, for instance, a standard learning method like cross-validation has as much justification as \emph{anti}-cross-validation (\citealp{zhu1996no}; \citealp[1359f]{Wol96nc}; \citeyear[378]{Wol21incol}).

Yet for many such standard learning algorithms we \emph{do} seem to have a justification. The field of machine learning theory is concerned with deriving mathematical learning guarantees, that purport to show that standard procedures, like minimizing empirical error on the training set, are better than other possible procedures, like \emph{maximizing} empirical error \citep{ShaBen14}. This raises a puzzle. How can there exist a learning theory at all, if the lesson of the \textsc{NFL} theorems is that learning algorithms can have no formal justification?

While this tension has been noted from the start \citep[1347]{Wol96nc}, existing explanations of the consistency of the \textsc{NFL} theorems with learning theory (e.g., \citealp[1368ff]{Wol96nc}, \citealp[202ff]{BouBouLug04inc}, \citealp[692ff]{LuxSch11incl}) are partial at best. In this paper, we investigate in detail the implications of the \textsc{NFL} results for the justification of machine learning algorithms. The main tool in our analysis is a distinction between two conceptions of learning algorithms, a distinction that has a parallel in the philosophical literature promoting a \emph{local} view of inductive inference. This is the distinction between a conception of learning algorithms as purely data-driven or \emph{data-only}, as instantiating functions that only take data, and a conception of learning algorithms as \emph{model-dependent}, as instantiating functions that, aside from input data, also ask for an input \emph{model}. %The model-dependent conception is the more usual and fruitful conception for the analysis of machine learning algorithms. 

We argue that the \textsc{NFL} theorems rely on the former, data-driven conception of learning algorithms; but that many standard learning methods, including empirical risk minimization and cross-validation, should not be viewed as such. By their specification, such algorithms take two inputs: data, and an explicitly formulated model or hypothesis class, which constitutes a choice of bias. What we can reasonably demand from such model-dependent algorithms is that they perform as well as possible \emph{relative} to any chosen model. Consequently, learning-theoretic guarantees are \emph{relative to} the instantiated models the algorithm can take, and it is in this form that there \emph{is} justification for standard learning algorithms. It is in this sense that learning theory allows one to say that empirical risk minimization is preferable to risk maximization, and that cross-validation is preferable to anti-cross-validation.

% , and there \emph{is} room for learning theory, which does allow one to  say that some learning algorithms work better than others. For example, cross-validation --- while certainly not optimal in all situations --- is definitely preferable over anti-cross-validation. 

%For instance, there exists a standard learning algorithm for neural networks, stochastic gradient descent with backpropagation, which must be supplied with a particular architecture, the actual network, that constitutes the model.
%PETER Dit is een beetje gevaarlijk voorbeeld omdat SGD niet zomaar
% een lokaal minimum vindt. Maar het is wel helder dus misschien toch
%maar zo laten staan.
%TOM to PETER: ik noem dit niet meer. Ik noem NN's nu alleen pas in de conclusie (en jouw stukje in de appendix).

This is all consistent with the valid lesson of the \textsc{NFL} results, namely that every data-only learning procedure must possess some inductive bias. Our point is that this lesson should not be taken as a stick to wield against any possible learning algorithm. On the contrary: in model-dependent learning algorithms, this lesson is accounted for from the start.

The plan of the paper is as follows. First, in section \ref{sec:equiv}, we introduce the original Wolpert-Schaffer results. Still granting here the data-only conception of learning algorithms, we dispute the results' interpretation that \emph{all algorithms are equivalent}. We discuss how this interpretation relies on an unmotivated assumption of a uniform distribution over possible learning situations, that can in fact be seen as an explicit assumption that learning is impossible. We advance the alternative statement that \emph{there is no universal data-only learning algorithm}. As instantiations of this statement, the \textsc{NFL} results illustrate and support the central insight in machine learning that every mechanical learning procedure, understood as a mapping from possible data to conclusions, must possess an inductive bias. %(We discuss the original results and the consistency with learning theory in more technical detail in appendix~\ref{app:orignfl}.)

Next, in section \ref{sec:nounivrule}, we develop the model-dependent conception of learning methods, and show how this conception makes room for a justification for standard learning methods that is consistent with the \textsc{NFL} results. We start by pointing out that discussions surrounding the \textsc{NFL} theorems share a questionable presupposition with Hume's original argument for inductive skepticism: the idea that the performance of our inductive methods must be grounded in a general postulate of the induction-friendliness of the world. We discuss philosophical work that denies the cogency of such a principle, and that advances a local view of induction. This leads us to a local view of learning algorithms: the model-dependent perspective, and the accompanying possibility of a model-relative learning-theoretic justification. We discuss this in more detail for Bayesian machine learning, empirical risk minimization, and cross-validation, making explicit why learning theory allows us to say, for instance, that cross-validation is more sensible than anti-cross-validation. %At the same time, we note some nuances to our argument, that we discuss in more detail in appendix \ref{app:details}. 
We conclude in section \ref{sec:concl}.

Finally, we provide two appendices that complement the main argument. In appendix~\ref{app:orignfl} we investigate the formal consistency of the original \textsc{NFL} results with learning theory, and in appendix~\ref{app:details} we list some important nuances to our discussion about model-dependent learning algorithms.

\section{All learning algorithms are equivalent?}\label{sec:equiv}
The first mentions in print of the ``no-free-lunch theorems'' of supervised learning are in  \citet[also see \citeyear{Wol95techrep}]{Wol95inc,Wol96nc},\footnote{Wolpert \citeyearpar[1343]{Wol96nc} attributes the term ``no-free-lunch theorems'' to the computer scientist D.\ Haussler. Wolpert and others also derived \textsc{NFL} theorems for mathematical optimization \citep{WolMac97ieeeec,hopep02jota}, which we do not discuss in this paper.} although an earlier version of the results already appeared in \citet{Wol92cs,Wol92techrep}. Around the same time, \citet{Sch94icml} presented a version of these results, with reference to Wolpert, as a ``conservation law for generalization performance.'' 

We start this section with presenting some basic versions of the Wolpert-Schaffer results, within a problem setting of prediction (sect.\ \ref{ssec:examplepred}), and within the original setting of classification (sect.\ \ref{ssec:exampleclassif}). Next, we discuss Wolpert's interpretation of his results that ``all learning algorithms are equivalent, on average.'' We discuss the results' concern with all possible learning algorithms vis-\`a-vis the traditional philosophical concern with ``inductive method,'' and note its restriction to data-only algorithms (sect.\ \ref{ssec:allalgos}). We then critically analyse Wolpert's equivalence claim and the underlying assumption of a uniform distribution over possible learning situations (sect.\ \ref{ssec:arequiv}). Finally, we advance the alternative \textsc{NFL} statement that there is no universal data-only learning algorithm (sect.\ \ref{ssec:nounivalgo}).

\subsection{Prediction}\label{ssec:examplepred}
\begin{figure}
  \begin{tikzpicture} [
    grow=up,
    level 1/.style={sibling distance = 6cm, font=\fontsize{10}{10}\selectfont, level distance = 1cm},
    level 2/.style={sibling distance = 3cm, font=\fontsize{8}{8}\selectfont},
    level 3/.style={sibling distance = 1.5cm, font=\fontsize{6}{6}\selectfont},
    every node/.style={circle,solid, draw=black,thin, minimum size = 0.5cm},
    emph/.style={edge from parent/.style={->,thick,draw}},
    norm/.style={edge from parent/.style={solid,black,thin,draw}}
    ]

    \begin{scope}[xshift=6cm]
    \node{$\varnothing$} 
    child[emph] { node {$\mathsf{T}$}
        child[emph] { node {$\mathsf{TT}$} 
			child[emph] { node {$\mathsf{TTT}$} }
            child[norm] { node {$\mathsf{TTN}$} }         
        }
        child[norm] { node {$\mathsf{TN}$}
            child[emph] { node {$\mathsf{TNT}$} }
            child[norm] { node {$\mathsf{TNN}$} } 
        }
    }
    child { node {$\mathsf{N}$}
        child[emph] { node {$\mathsf{NT}$} 
			child[emph] { node {$\mathsf{NTT}$} }
            child[norm] { node {$\mathsf{NTN}$} }         
        }
        child[norm] { node {$\mathsf{NN}$}
            child[emph] { node {$\mathsf{NNT}$} }
            child[norm] { node {$\mathsf{NNN}$} } 
        }
    }
    ;
    \end{scope}
  \end{tikzpicture}
  \caption{\small \textsc{NFL} for prediction. For any possible learning method (say, the method that always chooses $\mathsf{T}$, here represented by the arrows), there is one learning situation (path through the tree) with error 0 (follow the arrows), one with error 1 (never follow the arrow), and three situations each with error $1/3$ and $2/3$. Assigning each learning situation the same probability $1/8$, the algorithm's expected error is $1/2$.}\label{fig:pred}
\end{figure}
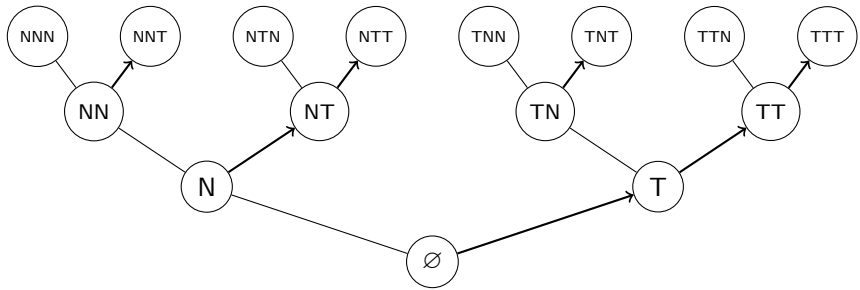

Imagine that every day we are given a bowl of oatmeal for breakfast. Every morning on waking up, before we have our breakfast, we seek to predict whether it will be tasty ($\mathsf{T}$) or not ($\mathsf{N}$), based only on when it was the days before. A \emph{learning algorithm} in this simple learning framework makes a guess whether the oats we are served today will be tasty, based on the data of the previous days. For a sequence of three days (see figure \ref{fig:pred}), there are in this scenario $2^3$ logically possible histories or \emph{learning situations} (of the form $\mathsf{TTT, TNT, NTT}$, \dots), and already $2^7$ possible learning algorithms (functions from $\{\emptyset, \mathsf{T,N, TT, NT, TN, NN}\}$ to $\{\mathsf{T,N}\}$). Let an algorithm's \emph{error} be the ratio, among all predictions, of those predictions that are \emph{incorrect} (e.g., a prediction of $\mathsf{T}$ and then obtaining $\mathsf{N}$). Then a no-free-lunch statement in this scenario is that \emph{for each possible level of error, every learning algorithm suffers this error in equally many possible learning situations}. Namely, one can verify that every single algorithm predicts perfectly (has error 0) in exactly one possible learning situation, predicts maximally badly (error 1) in exactly one other possible situation, suffers error $1/3$ in three possible learning situations, and error $2/3$ in the remaining three.\footnote{A similar example is given by \citet[551f]{For99mm}.} 

Note that in thus counting learning situations and comparing these counts, we treat all possible learning situations on a par. Another way of doing this is to assume a \emph{uniform probability distribution} on all possible learning situations, that is, a distribution that assigns the same probability to each of the finitely many possible learning situations. Then the above \textsc{NFL} result can be restated as the observation that, under the uniform distribution on learning situations, \emph{every learning algorithm has the same expected error} of exactly 1/2. That is, every learning algorithm can be expected to do no better (or worse) than random guessing.\footnote{This
  statement generalizes to learning algorithms that issue probabilistic predictions, and the error of a single prediction  $p \in (0,1)$ is given by $p$ in case of outcome $\mathsf{N}$, and $1-p$ otherwise. Then the more cautious a learning algorithm (the closer its predictions are to $1/2$), the smaller the number of histories on which it attains either very low or high error, but this evens out in such a way that every learning algorithm still has an expected error of $1/2$ (see \citealp[7ff]{Sch21epi}).}

\subsection{Classification}\label{ssec:exampleclassif}
The original Wolpert-Schaffer results were derived in a problem setting more standard in machine learning theory, the setting of \emph{classification}. We first discuss the simplified setting of \emph{non-stochastic} classification (sect.\ \ref{sssec:exampleclassifns}), before we turn to the more general setting of \emph{stochastic} classification (sect.\ \ref{sssec:exampleclassifs}). 

%TOM2: sectie over classificatie was wel erg lang geworden, dus opgedeeld in twee subsecties
\subsubsection{Non-stochastic classification}\label{sssec:exampleclassifns}
Imagine we want to learn to successfully classify whether a bowl of oats will be tasty or not, based on three different features we can determine before trying it: its temperature, its color, and its smell. Formally, supposing that these attributes are binary (either hot or cold, either bright or dull, either reeking or not), every \emph{instance} of a bowl of oats can be represented by a length-three attribute vector of binary (write 0 or 1) components. This gives a total of eight ($2^3$) different possible instances, collected in the \emph{domain set} $\mathcal{X}=\{0,1\}^3$. A \emph{classifier} is a function $f: \mathcal{X} \rightarrow \mathcal{Y} $ from the possible instances to their \emph{labels} (tasty or not), collected in the \emph{label set} $\mathcal{Y} = \{\mathsf{T,N}\}$. Supposing that the true labels are indeed fully determined by the attributes, the possible learning situations---possible \emph{true} labelings of all instances of oats---are given by the possible classifiers.  A \emph{learning algorithm} $A$ maps a sample $S=(x_1,y_1), \dots, (x_n, y_n)$ of \emph{training data}, pairs of instances and true labels, to a particular classifier $f$. 

We are now interested in a learning algorithm's \emph{generalization error} $L_{\overline{S}}(A(S))$: given some training sample $S$, how accurate is the classifier $f=A(S)$ selected by $A$ on the instances that lie outside of $S$? Suppose the training data includes six of the total number of eight different possible instances of oats, determining the true tastiness labels for these six instances (see table~\ref{fig:classf}). There are four possible ways of classifying the two unseen instances, or four remaining possible learning situations $f^*$. Each possible learning algorithm selects a particular classifier in response to the training data, which classifies the two unseen instances in one of the four possible ways. That means that each possible learning algorithm (selected classifier $f$) has the same generalization error (ratio of incorrectly classified unseen instances over all unseen instances: either 0, 0.5, or 1) in the same number (one, two, one) of still possible learning situations $f^*$.\footnote{Implicit 
  here (and elsewhere in our presentation) is the use of a particular function to measure error or \emph{loss}, the (standard) 0/1 loss function. The  \textsc{NFL} theorems as stated do not necessarily go through for other loss functions \citep{Wol96nc2}.}

\begin{table}[h!] 
\begin{center}
\begin{tabular}{ c | c c c|c c c c c c c c c c } 
 \ & temp. & color & smell & \multicolumn{9}{c}{tastiness, according to} \\
 &  &  & & $\hat{f}$ & $f^*_1$  & $f^*_2$ & $f^*_3$ & $f^*_4$ & $f^*_5$ & $f^*_6$ & $f^*_7$ & \dots & $f^*_{256}$ \\  
 \hline\hline 
 \multirow{6}{*}{\parbox{1.5cm}{training sample $S$}} & 0 & 0 & 0 & $\mathsf{N}$ & $\mathsf{N}$ & $\mathsf{N}$ & $\mathsf{N}$ & $\mathsf{N}$ & $\mathsf{N}$ & $\mathsf{N}$ & $\mathsf{N}$ &  \dots & $\mathsf{T}$ \\ 
 & 0 & 0 & 1 &  $\mathsf{N}$ & $\mathsf{N}$ & $\mathsf{N}$ & $\mathsf{N}$ & $\mathsf{N}$ & $\mathsf{N}$ & $\mathsf{N}$ & $\mathsf{N}$ &  \dots & $\mathsf{T}$ \\ 
 & 0 & 1 & 0 & $\mathsf{N}$ & $\mathsf{N}$ & $\mathsf{N}$ & $\mathsf{N}$ & $\mathsf{N}$ & $\mathsf{N}$ & $\mathsf{N}$ & $\mathsf{N}$ &  \dots & $\mathsf{T}$ \\ 
 & 1 & 0 & 0 & $\mathsf{N}$ & $\mathsf{N}$ & $\mathsf{N}$ & $\mathsf{N}$ & $\mathsf{N}$ & $\mathsf{N}$ & $\mathsf{N}$ & $\mathsf{N}$ &  \dots & $\mathsf{T}$ \\ 
 & 0 & 1 & 1 & $\mathsf{N}$ & $\mathsf{N}$ & $\mathsf{N}$ & $\mathsf{N}$ & $\mathsf{N}$ & $\mathsf{N}$ & $\mathsf{N}$ & $\mathsf{N}$ &  \dots & $\mathsf{T}$ \\ 
 & 1 & 0 & 1 & $\mathsf{N}$ & $\mathsf{N}$ & $\mathsf{N}$ & $\mathsf{N}$ & $\mathsf{N}$ & $\mathsf{T}$ & $\mathsf{T}$ & $\mathsf{T}$ & \dots & $\mathsf{T}$ \\
 \hline
 \multirow{2}{*}{\parbox{1.5cm}{unseen instances}} & 1 & 1 & 0 & $\mathsf{N}$ &$\mathsf{N}$ & $\mathsf{N}$ & $\mathsf{T}$ & $\mathsf{T}$ & $\mathsf{N}$ & $\mathsf{N}$ & $\mathsf{T}$ & \dots & $\mathsf{T}$ \\ 
 & 1 & 1 & 1 & $\mathsf{T}$ & $\mathsf{N}$ & $\mathsf{T}$ & $\mathsf{N}$ & $\mathsf{T}$ & $\mathsf{N}$ & $\mathsf{T}$ & $\mathsf{N}$ & \dots & $\mathsf{T}$ \\ 
\end{tabular}
\captionof{figure}{\small \textsc{NFL} for nonstochastic classification. For any learning algorithm $A$, any non-exhaustive training sample $S$ (here of size six) and any possible labeling of $S$ (say, all $\mathsf{N}$, leading $A$ to output classifier $\hat{f}$), there is the same number (here, four) of remaining possible learning situations (here, the classifiers $f^*_1$ to $f^*_4$) that each label the (here, two) remaining instances differently. (Table adapted from \citealp{GirPro05icml}.)}\label{fig:classf}
\end{center}
\end{table}

Alternatively, we can put things again in terms of a uniform distribution $\mathcal{U}$ over all possible learning situations. So for this specific sample $S$ of instances and labels, we have that uniformly averaged over the four remaining possible learning situations, the error of each learning algorithm is equal to 1/2. More generally, we can consider the same sample $S_X$ stripped of its labels, and move the averaging to the front, so to speak, to cover how the possible $f^*$ (now \emph{all} possible $f^*$) assign labels to $S_X$, and how the algorithm fares for the resulting $S=S_X \times f^*(S_X)$ of instances and labels. But since for any four learning situations that label $S_X$ in an identical way, an algorithm's average generalization error is 1/2, it remains 1/2 when averaged this way over \emph{all} learning situations; and this reasoning goes through for any non-exhaustive $S_X \subsetneq \mathcal{X}$. Thus we arrive at the statement that for any non-exhaustive training sample $S_X$ of instances every learning algorithm $A$ has expected generalization error $\Expec_{f^* \sim \mathcal{U}}\left[ L_{\overline{S}}(A(S)) \right] = 1/2 $.\footnote{Similar 
  illustrations of the \textsc{NFL} theorems are given by \citet[454ff]{DudHarSto01}; \citet[9f]{GirPro05icml}; \citet[1900f]{Bar11nc}; \citet[720ff]{OrtLei11incl}; \citet[224ff]{LatHut13inproc}. A precursor to this variant is the ``theorem of the ugly duckling'' due to \citet[376ff]{Wat69}.}

\subsubsection{Stochastic classification}\label{sssec:exampleclassifs}
An additional refinement in the standard framework for classification (see \citealp{ShaBen14}) is that the true connection between instances and labels can itself be stochastic. Moreover, we assume some unknown probability distribution for the drawing of instances. Thus a learning situation is given by a distribution $\mathcal{D}$ over pairs of instances and labels.% in $\mathcal{X} \times \mathcal{Y}$; we imagine we repeatedly draw an object $x$ from the marginal distribution $\mathcal{D}_\mathcal{X}$ over $\mathcal{X}$, that is assigned a label via the conditional distribution $\mathcal{D}(Y \mid x)$.
\footnote{The 
  problem setting presupposed by Wolpert (his ``extended Bayesian formalism,'' also see footnote \ref{fn:ebf}) is more general still, in that both the classifier and the learning algorithm are stochastic: a classifier, like the true connection between instances and labels, is a $\mathcal{Y}$-conditional distribution over $\mathcal{X}$, and a learning algorithm is a distribution over classifiers. This additional generality does not affect the \textsc{NFL} statement in this section (cf.\ \citealp[473; 478f]{RaoGorSpe95icml}).}  
  
We now also measure generalization error in expectation over drawing an instance from $\mathcal{D}$: we shall call this the \emph{risk}. But we have a choice here: do we take the expectation over all over $\mathcal{X}$, so including instances that were already in the training set, or do we discard the latter? Wolpert's ``off-training-set'' (\textsc{ots}) risk, write $L_{\mathcal{D} \setminus S}(A(S))$, explicitly discounts already seen instances. He actually departs here from most of learning theory, where the error is standardly evaluated over all instances. We shall follow Wolpert in calling the latter quantity ``i.i.d.''\ (\textsc{iid}) risk, write $L_\mathcal{D}(A(S))$. 
%PETER2 Heel goed dat je dat explicieter hebt gemaakt. Maar ik denk dat we nu all the way moeten gaan en ook even expliciete formules moeten geven, ook al is dit misschien niet de gewoonte in een filosofie paper. Deze foutmaten zijn nl essentieel voor wat gaat komen en er komt een hoop af op de lezer in sectie 2.2. dus we moeten verwarring voorkomen.
%Dus hier zijn expliciete formules. Ik heb hier gelijk impliciet uitgeled wat we met 0/1 loss precies bedoelen.
%Wel ziet er dit nu misschien wat al te technisch uit misschien. Maar helemaal zonder deze formules lijkt me ook erg lastig
%TOM2: Dat lijkt me prima, ik heb de formules wel even apart gezet voor betere leesbaarheid.
Formally, for given sample $S= (x_1,y_1), \ldots, (x_n,y_n)$, $L_\mathcal{D}(A(S))$ is the probability, under $\mathcal{D}$, that an independently sampled example $(X,Y)$ has $f(X) \neq Y$, where $f= A(S)$ is the classifier output by algorithm $A$ on input $S$. This can also be written as 
\begin{equation}
L_\mathcal{D}(A(S)) = {\bf E}_{(X,Y) \sim \mathcal{D}} [|Y-A(S)(X)|],
\end{equation}
that is, the expected $0/1$-error. In contrast,  $L_{\mathcal{D} \setminus S}(A(S))$
is the probability that $f(X) \neq Y$, with $f= A(S)$, conditional on $(X,Y) \not \in \{(x_1,y_1), \ldots, (x_n,y_n) \}$.  This can also be written as 
\begin{equation}L_{\mathcal{D} \setminus S}(A(S)) = {\bf E}_{(X,Y) \sim \mathcal{D}}\left[|Y-A(S)(X)| \; \mid  X,Y \not \in \{(x_1,y_1), \ldots, (x_n,y_n) \}\right].
\end{equation}

A central claim in Wolpert's works is that \textsc{ots} risk is a more natural measure of \emph{generalization} performance than \textsc{iid} risk (\citeyear[1345ff]{Wol96nc}; \citeyear[25ff]{Wol02inc}). Note that it is certainly more similar to the generalization error in the previous nonstochastic case (where the labels of already seen instances are conclusively learned). But this does not make it clearly better in the stochastic case, where there is still an estimation problem even for already seen instances. We discuss the relation between the two notions (and the relevance of their difference in the context of the consistency of the \textsc{NFL} results with positive results in learning theory) in more detail in appendix~\ref{ssec:appwolots}, and in the following always make clear what risk we mean.

Abstracting away from the oatmeal classification example, suppose instances are given by some finite-length set of features that can take a finite number of values, so that there is a (possibly huge yet) finite number $m$ of possible instances. %Suppose we draw from $\mathcal{D}$ a training set of $n < m$ labeled instances, leaving $k \geq m-n$ (accounting for the possibility of duplicates in the training data) unseen instances. 
Given some training set $S$ of $n$ labeled instances, consider again any single unseen instance $x$. For each learning algorithm (selected classifier, assigning label $y$ to $x$), there is a possible learning situation $\mathcal{D}$ in which the classifier's risk on this particular $x$ is 0 (namely, a $\mathcal{D}$ that assigns probability 1 to label $y$, conditional on instance $x$). Likewise, there is a possible learning situation $\mathcal{D}$ in which the classifier's risk on this particular $x$ is 1. Indeed, for each value in the unit interval there is a possible learning situation in which the classifier has \emph{that} risk on $x$, as well as a counterpoint situation where the classifier has \emph{one minus} that risk on $x$. The intuition that these risks all even out finds again a precise expression under the assumption of an (in this case, \emph{continuous)} uniform distribution $\mathcal{U}$ over all learning situations---in this case, a uniform distribution\footnote{In 
	general, ``uniform distributions'' over general spaces are ill-defined or (if the spaces are noncompact) do not even exist: in appendix~\ref{ssec:appwolcons} we show how in the current setting, an unambiguous definition is possible.} \emph{over distributions}. Thus for any given set of training data, for any learning algorithm, the selected classifier's $\mathcal{U}$-expected risk on any single unseen instance is 0.5. This concerns a specific unseen instance, given some specific set of training data. But, crucially, we can again move the expectations to the front, to range over the whole process of drawing training data and measuring risk.\footnote{See 
the exhaustive reconstruction by \citet{RaoGorSpe95icml} for details. An illustration similar to ours is given by \citet[693f]{LuxSch11incl}.} 
  In this way we reach the statement of the \textsc{NFL} theorem, or the conservation law of generalization performance: every learning algorithm $A$, for any sample size $n<m$, has the same expected \textsc{ots} risk 
 $\Expec_{\mathcal{D}\sim \mathcal{U}, S \sim \mathcal{D}^n}
\left[ L_{\mathcal{D} \setminus S}(A(S))\right]=1/2$.\footnote{The original statement by Wolpert and also Schaffer is still slightly different from the statement we give here. They actually take apart the marginal distribution $\mathcal{D}(X)$ that generates the instances and the conditional distribution $\mathcal{D}(Y \mid X)$ that labels the instances, allow the former to be any distribution, and let the uniform distribution $\mathcal{U}$ only range over the latter.
We state this precisely in appendix~\ref{ssec:appwolcons}.}

%PETER2 BELANGRIJK... hier klopt nog iets niet helemaal. Ik heb Wolpert nagekeken en volgens mij hoef je eigenlijk niets over de verdeling over X
%aan te nemen voor de NFL results, je moet alleen een uniforme prior hebben op de  verdeling van Y|X=x, geconditioneerd op elke x. 
%Maar D is een verdeling *over Y EN x*, en hier doe je alsof die hele 'joint'verdeling zelf een uniforme prior over zich heen heeft.
%Maar het is te lastig om dit correct en toch niet heel ingewikkeld op te schrijven, dus ik stel dat maar uit tot de appendix
%TOM2: ja, dat klopt, Wolpert en ook Schaffer nemen dat niet aan. In de gedetailleerde reconstructie van Rao et al. (1995), waar ik ook veel van gebruikt maakte, doen ze dat trouwens wel, vandaar dat ik dacht dat deze vereenvoudiging niet zoveel uitmaakte.
%TOM2: Ik heb dingen een klein beetje herschreven en een voetnoot toegevoegd.
%Here intuitively, $\mathcal{D} \sim \mathcal{U}$ means that $\mathcal{D}$ `is itself uniformly distributed'. The precise meaning of this statement is given in Appendix~\ref{ssec:appwolots}.  

\subsection{All learning algorithms \dots}\label{ssec:allalgos}
We presented some versions of the Wolpert-Schaffer results, leading up to what is essentially the original form. But already the first example in the framework of prediction brings out an important characteristic of the \textsc{NFL} theorems:  their concern, for the given learning problem, with \emph{all possible learning algorithms}, understood as mappings from data to conclusions.\footnote{\label{fn:algofn}We 
  speak of ``algorithms,'' following the custom in discussions surrounding the \textsc{NFL} results, even if it is perhaps better to speak of (for instance) learning \emph{functions}. In reality, the same ``algorithm'' (map from data to conclusions) can be implemented---or indeed approximated---by many different (say, more or less computationally efficient) algorithms.}

There is, to begin with, no special regard for particular subclasses of learning algorithms, say those that we would intuitively call ``inductive'' (or indeed \emph{learning} algorithms). In the prediction setting, for example, an ``inductive'' function that extrapolates the past data $\mathsf{NN}$ to the prediction $\mathsf{N}$ is no less a learning algorithm than an ``anti-inductive'' function that extrapolates data $\mathsf{NN}$ to prediction $\mathsf{T}$, or indeed than the ``learning-resistant'' constant function that outputs $\mathsf{T}$ no matter what. As such, the \textsc{NFL} theorems can be seen to simply bypass the main companion problem to that of \emph{justifying} induction: the problem of \emph{specifying} or \emph{describing} what actually constitutes inductive method or methods (see \citealp[7ff]{Lip04}).\footnote{\label{fn:newriddle}In
  particular, we can understand the infamous riddle of induction due to Goodman \citeyearpar{Goo54} as an expression of this problem of description. Suppose, on a minimal understanding of induction as ``extrapolating the pattern from the past into the future,'' that we somehow were assured that inductive inference \emph{is} justified: then, Goodman writes, we still would not know how to actually \emph{do} induction. There are always multiple patterns we can find in the past, hence always multiple (and inconsistent) ways we can extrapolate these. Rather than following the route of attempting to specify which of the many possible extrapolations constitutes a proper inductive inference (the route Goodman himself took with his notion of \emph{projectability}), we can remain agnostic and refrain from excluding \emph{any} formal extrapolation rule: and indeed the \textsc{NFL} theorems apply to all of them.} 

That said, when it comes to the assessment of the results' implications, it seems there is only a small subset of all logically possible algorithms that we are really interested in. % When it comes to the results' implications for the justification for learning algorithms, what we are really interested in
These are the \emph{algorithms that are actually used}. There is a limited number of standard algorithms developed and analyzed in machine learning, generic algorithms that are employed in a wide variety of different domains. Naturally enough, the motivating discussions in Wolpert's writings focus on the ramifications of his results for the justification for \emph{these} algorithms. We will discuss the justificatory implications of the \textsc{NFL} in detail in sect.\ \ref{sec:nounivrule} below.

While the ``all possible'' in the \textsc{NFL} results' characteristic concern with \emph{all possible learning algorithms} can be seen as a useful generality in the results' scope, there is also an % inevitable but still 
important sense in which this scope is limited. This has to do with the restriction to ``learning algorithms,'' understood as well-defined mappings from data to conclusions. The \textsc{NFL} results apply to formal learning rules that fully specify what conclusion follows which observed data. They clearly do not apply to a non-algorithmic conception of inductive method(s) that involves irreducibly informal factors (like, perhaps, everyday human and even scientific reasoning). But they do not even apply to a conception of learning methods as taking for input other (context-dependent) elements: the \textsc{NFL} results apply to a conception of learning algorithms as purely data-driven or \emph{data-only}. We will also return to and expand on this point in sect.\ \ref{sec:nounivrule} below.
 
\subsection{\dots \ are equivalent?}\label{ssec:arequiv}
The interpretation that Wolpert attached to his formal results, and that we went along with in our presentation, is that ``for any two learning algorithms A and B \dots \ there are just as many situations (appropriately weighted) in which algorithm A is superior to algorithm B as vice versa'' \citep[1360]{Wol96nc}, or that ``all algorithms are equivalent, on average'' (\citealp[129]{Wol95inc}; \citeyear[35]{Wol02inc}). The obvious worry about the significance of the \textsc{NFL} theorems concerns the qualifiers ``appropriately weighted'' and ``on average'' in these statements: that is, the presupposition of a uniform distribution on learning situations. This is indeed what the immediate responses in the literature focused on. 

Perhaps the main criticism is that a uniform distribution is really a \emph{worst-case} assumption for the purpose of learning. The ``rational reconstruction'' by \citet{RaoGorSpe95icml} shows that Schaffer's conservation law of generalization performance is equivalent to the (trivial) statement that for any unseen example, both possible classifications result in a generalization error of 0.5, \emph{if} we measure the latter by uniformly averaging over both possible true classes. On a more conceptual level, this procedure of uniformly averaging corresponds to assuming that however many examples we have seen, we cannot have \emph{learned} anything: the best guess for the label of any new example will always still be fifty-fifty. Thus these authors conclude that ``the uniform concept distribution \dots \ in which every possible classification of unseen cases is equally likely \dots \ is the definition of a uniformly random universe, in which learning is impossible'' (ibid, 475).%\textsuperscript{,}
\footnote{In
   appendix~\ref{app:orignfl}, where we discuss the formal consistency of the original Wolpert-Schaffer results with learning theory, we bring out the same point in yet another way.} 
   Obviously the \textsc{NFL} theorems cannot be said to hold much significance if we understand them as the observation that every learning algorithm is equivalent \emph{in a universe where learning is impossible}.\footnote{Also 
  see the discussion by \citet{Sch17pos} of the \textsc{NFL} theorems in the setting of prediction, with the corresponding ``state-uniform'' prior that assigns equal probability to every same-length sequence of outcomes. Schurz connects this to Carnap's discussion of the corresponding confirmation function $\mathfrak{c}\dagger$, ``tantamount to the principle never to let our past experiences influence our expectations of the future \dots \ in striking contradiction to the basic principle of all inductive reasoning'' \citep[565]{Car50}, and also points out that the corresponding uniform measure on the space of all infinite sequences assigns probability 1 to infinite sequences (1) having a limiting relative frequency 1/2 and (2) being incomputable. Schurz concludes that ``proponents of a state-uniform prior distribution are strongly biased: they are a priori certain that the world is irregular so that induction cannot have any chance'' \citeyearpar[834]{Sch17pos}. The point that certain uniform distributions lead to ``unlearnability'' goes back at least to \citet{Boole1854laws}: if we assume that each ball in a bag has an equal probability of being black or white, he writes, then ``past experience [of drawing with replacement] does not in this case affect future experience'' (ibid., 372).}

It has been suggested that this particular criticism can be countered by the observation that a uniform distribution is not a necessary condition for \textsc{NFL} theorems to go through (e.g., \citealp[10]{GirPro05icml}). Rao et al.\ \citeyearpar[475ff]{RaoGorSpe95icml} show that generalization performance is conserved under a wider class of distributions; and indeed Wolpert \citeyearpar[1361f]{Wol96nc} also already gives ``extensions for nonuniform averaging.'' But as long as the results do not extend to \emph{all} distributions (and they do not: there is a certain symmetry that must be retained, \citealp[477]{RaoGorSpe95icml}), the worry remains that the \textsc{NFL} results are simply an expression of the induction-hostileness of the presupposed weighing distribution. 

Wolpert was aware of this perspective on his results.\footnote{His discussion of the intuition behind his results in (\citeyear[133f]{Wol96inc}; \citeyear[38ff]{Wol02inc}) is in fact very similar to the analysis of Rao et al.\ in tracing the results back to the assumption of fully random labels of unseen instances.} In \citeyearpar{Wol92cs}, he himself refers to the assumption of a ``maximum-entropy universe''; the way he puts his point there is that ``[s]ince such a universe cannot be ruled  out on  an a priori basis, it is theoretically  impossible  to  come  to  any  conclusions  about how to generalize using \emph{only} a priori reasoning.'' But the statement that it is a priori \emph{possible} that there are (in expectation) no distinctions between learning algorithms is weaker than the categorical statement that there \emph{are} (in expectation) no a priori distinctions between learning algorithms, the claim of the later paper \citeyearpar{Wol96nc}.

In this paper (ibid., 1362ff), Wolpert actually argues that the uniform distribution does have a preferred status. %[he writes that it is not so surprising that there is no optimal algorithm, but that it is interesting that we cannot even say that some algorithms are usually better than others (if we weigh targets as to the algo's success).] 
He starts by allowing that if we change the weighing of learning situations, then there could arise ``a priori distinctions'' between learning algorithms. However, he continues, ``a priori'' such a change of weighing could just as well favor algorithm A as B: ``[a]ccordingly, claims that `in the real world [the distribution over learning situations] is not  uniform, so the NFL results do not apply to my favorite learning  algorithm' are misguided at best'' (ibid., 1363). Indeed, he points at results in the same paper regarding averages over \emph{prior distributions} over learning situations, with the interpretation that there are as many \emph{priors} for which A is superior to B as the other way around. From this perspective, ``uniform distributions over targets are not an atypical, pathological case \dots \ [r]ather they and their associated results are the average case (!)'' (ibid.).

This jump to a higher level is clearly inconclusive: we can restate the same worry at \emph{that} level.\footnote{Elsewhere (\citeyear[footnote 3]{Wol95inc}; \citeyear[footnote 4]{Wol02inc}), Wolpert presents the situation rather in terms of the critic of the uniform distribution attempting to ``jump a level'' in questioning the uniform distribution on priors, ``arguing that some [prior distributions over learning situations] are `more likely' than others''---but to no avail, ``the math responds the same way as it did to the [lower-level] objection.'' This is a reiteration of the dialectical move we criticize next.} Most remarkable, however, is Wolpert's dialectical move of turning the table on the critic: ``the burden is on the user of a particular learning algorithm. Unless they can somehow show that [the true prior] is one of the ones for which their algorithm does better than random \dots \ they cannot claim to have any formal justification for their learning algorithm'' (ibid.).

Curiously, responses in the computer science literature critical of the significance of Wolpert's results have essentially \emph{followed} him here. Rao et al., after discussing how \textsc{NFL} theorems must depend on a symmetrical prior, conjecture that ``our  world has strong regularities, rather than  being nearly random. However,  only time and further testing of physical theories can refine our understanding of the nature of our universe [and] might lead  to a  reasonable estimate of [the true prior] in our world'' \citeyearpar[477]{RaoGorSpe95icml}. Giraud-Carrier and Provost emphatically set forth as an implicit yet generally accepted ``weak assumption of machine learning'' that ``the process that presents us with learning problems \dots \ induces a non-uniform probability distribution [over learning situations]'' \citeyearpar[11]{GirPro05icml}.\footnote{The accompanying \emph{strong} assumption of machine learning is that this distribution is actually ``explicitly or implicitly known, at least to a useful approximation'' \citep[11]{GirPro05icml}.} But this Wolpert would not disagree with: he writes himself that a nonuniform distribution ``is why some algorithms tend to perform better than others \emph{in the real world}'' \citep[1361, emphasis ours]{Wol96nc}.\footnote{In 
	the introduction of his paper, Wolpert writes that ``[i]t cannot be emphasized enough that no claim is being made \dots \ that all algorithms are equivalent \emph{in practice}, in the real world \dots \ The sole concern of this paper is what can(not) be formally verified about the utility of various learning algorithms if one makes no assumptions concerning targets'' \citeyearpar[1344]{Wol96nc}. Also see \citet[61]{Wol92cs}.} 
	The point is to give a ``formal justification'' for believing in any such distribution. Indeed, if we seek to criticize the assumption of a uniform distribution in Wolpert's claim that all algorithms are a priori equivalent \emph{by postulating a different, nonuniform, distribution}, then we better provide a justification for postulating \emph{that} distribution. The result is that we find ourselves in a corner, because it is not clear where to look for such a justification. What we should have done, of course, is to insist that Wolpert justify \emph{his} assumption.

In fact, a more fundamental reply is to demand a reason for postulating \emph{any} prior distribution over learning situations. Doing so is a formal requirement in Wolpert's ``extended Bayesian formalism'' (unlike in the conventional classification framework); but that merely shows that the framework is constraining in a way that we may find inpalatable.\footnote{\label{fn:ebf}In
  Wolpert's \textsc{EBF}, one defines a probability distribution $P$ that ranges over ``target functions'' $f$ as well as ``hypotheses'' $h$, where the latter stand for the learning algorithm's possible guesses for the true target $f$. So $P(f)$ and $P(f\mid d)$ represent the ``true'' or ``objective'' priors and posteriors over targets, and $P(h\mid d)$ the learning algorithm. The need to thus specify a prior over targets $f$ deviates strongly from the conventional learning theory framework, where it is only assumed that there is some unknown distribution governing instances and labels.}\textsuperscript{,}\footnote{The 
  name of Wolpert's framework derives from its aim to generalize the conventional Bayesian framework where ``there is no direct  analogue to $P(h|d)$ \dots \ Viewed  another way, [it] has $P(h|d)$ pre-fixed, to be the `Bayes-optimal' $P(f|d)$'' \citep[122]{Wol95inc}. Thus, under the Bayesian interpretation of probability as degree of belief, ``you automatically know $P(f)$ exactly.'' But ``a `truth' $f$ and a guess $h$ are different objects \dots \  a formal statement connecting $P(f|d)$ and $P(h|d)$ corresponds to an extra assumption not demanded by the mathematics.'' \citeyearpar[83f, notation aligned with the previous]{Wol96inc}. This is, to put it mildly, an idiosyncratic rendering of the Bayesian approach. Rather than conflating, absurdly, an epistemic and an ontic interpretation of the prior $P(f)$, a Bayesian would stay clear of the latter---that is neither demanded by mathematics, but presupposed by Wolpert.}
  Indeed, it is not at all clear what it is supposed to \emph{mean} to assign probabilities to possible learning situations. An epistemic interpretation, as some (ideally rational) agent's degrees of belief, is perhaps the easiest to make sense of, but immediately throws us back to the justification for any specific choice of prior distribution: in particular, the idea of a uniform distribution as an objective-logical ``indifference prior'' has long been abandoned by philosophers and statisticians alike as a viable option (see, e.g., \citealp[293ff]{Fra89}; \citealp{Zab16inc}).
%PETER to TOM: zou goed zijn om hier nog 1 a 2 citaten toe te voegen.
%Dit is wel hoe velen erover denken (indifference) dus hoe duidelijker
%we kunnen maken dat het nergens op slaat, hoe beter. 
%TOM to PETER: heb jij toevallig wat goede referenties uit de statistiekhoek bij de hand? wat ik hier zeg is eigenlijk al een beetje te sterk als het om filosofie gaat, waar een bepaalde afgebakende vorm van indifference nog steeds wel onderdeel uitmaakt van actieve programma's. maar dit is niet eens van heel groot belang in de context van Wolperts argument, omdat (zoals ik vervolgens schrijf) hij niet een epistemische maar daadwerkelijk een objectief-fysische lezing van die prior in gedachten heeft, wat werkelijk geen fundament heeft.
%PETER2: goed punt. Ik heb onze oude vriend Seidenfeld (in een paper dat vooral door statistici geciteerd wordt) erbij gezet, denk dat het zo wel voldoende is
This is, in any case, not what Wolpert appears to have in mind: the suggestion is rather that we should think of these probabilities as objective-\emph{physical}, as chances.\footnote{See especially \citet[84]{Wol96inc}: ``if we had sufficient knowledge of the laws of physics (in particular, the boundary conditions of the universe) and of the (resultant) laws of human psychology, and if we were sufficiently competent to perform the appropriate quantum mechanical calculatations, then we might say that we could calculate [the distribution of learning situations] exactly.'' Again, the \emph{objective} interpretation of distributions over learning situations, complementary to the \emph{learning algorithm's} distribution over ``hypotheses,'' is central to Wolpert's framework.} But in the absence of a fuller account of the nature of these chances we do not see much reason for going along with the idea that the universe is governed by some objective distribution generating learning situations---let alone that this distribution should be uniform.

In sum, it would be granting Wolpert too much to accept that it is on us to show, contra his equivalence claim, that some algorithms are generally better than others. (We do not even need to think that ``generally'' is a qualifier that can be made meaningful here.) The burden is rather on Wolpert to justify the presuppositions that back his claim, in particular the assumption of a uniform distribution on learning situations, and this he has not done.  

\subsection{There is no universal data-only learning algorithm}\label{ssec:nounivalgo}
We can, however, formulate a weaker variation of the \textsc{NFL} results, a statement that is implied by the original but that does away with the uniformity assumption. In stating it, we also make explicit the observation from sect.\ \ref{ssec:allalgos} that we are still talking of \emph{data-only} algorithms, functions from data to conclusions:
\begin{equation}
  \tag{$*$}\label{eq:NFL}
  \parbox{\dimexpr\linewidth-4em}{%
    \strut
    For any data-only learning algorithm, there exists a learning situation in which this algorithm does not perform well, while in this same situation another data-only learning algorithm does perform well.
    \strut
  }
\end{equation}

%PETER to TOM: HIER is het belangrijk dat de casual reader direct ziet dat we  het over purely data driven learning algorithms hebben. Het zou echt fijn zijn als we deze in het hele paper consequent met een qualifier aanduiden (bijv. data-only learning algorithms vs model-dependent learning algoritms; of Type-I en Type-II learning algorithms 
% want iemand die het paper kort scant kan wat hier staat niet rijmen
% met sectie 3, denk ik 
%TOM to PETER: done. ik heb het statement ook wat uit laten springen en gelabeld voor latere referenties, en een stukje toegevoegd over leerraamwerk-afhankelijkheid.

In other words, %every data-only algorithm fails to perform well in some possible learning situation in which another data-only algorithm does perform well. In different terms still,
 there is no single data-only learning algorithm that performs well \emph{whenever some} data-only algorithm performs well: there is no \emph{universal} data-only learning algorithm.

Note right away that the truth of any instantiation of this statement depends on the learning problem in question, including the possible methods and the adopted notion of good performance. It is not too hard to come up with (artificial) learning problems for which the statement is false (e.g., a problem that is formulated such that the possible learning situations explicitly accommodate a particular learning method).\footnote{In 
  fact, Wolpert's own framework provides another example, if we see the uniformity assumption as \emph{part of the formulated learning problem}. This renders the problem trivial because there is only one possible ``truth'' or learning situation (the ``no-learning'' truth where the correct classifications are uniformly random) so that the statement must be false (in this case, because all algorithms are equally good, in terms of expected risk, in this one situation, hence ``universal'').} 
  The statement is relevant insofar it holds for problems within most standard learning frameworks and natural measures of good performance.

%In the initial non-stochastic classification setting we considered in section~\ref{ssec:exampleclassif}, we retrieve this statement if we make ``good performance'' precise as (for instance) ``having generalization accuracy strictly greater than 1/2.'' Namely, for every possible learning algorithm, for any training set of instances, there exists a learning situation (true classifier) such that the algorithm's selected classifier has generalization accuracy below 1/2 (indeed, equal to \emph{zero}, for true labels of the unseen instances orthogonal to the guesses of the classifier), while \emph{another} learning algorithm (for instance, the one that has selected this true classifier) has  generalization accuracy above 1/2 (indeed, \emph{one}).

For instance, we retrieve this statement from the original Wolpert-Schaffer result if we drop the uniformity assumption and make ``good performance'' precise as (say) ``having expected risk strictly smaller than 1/2.''  Namely, for every learning algorithm $A_1$, for any sample size $n$, there exists a learning situation $\mathcal{D}$ such that the algorithm has expected \textsc{ots} risk $\Expec_{S \sim \mathcal{D}^n}\left[ L_{\mathcal{D}\setminus S}(A_1(S)) \right]$ at least 1/2, while \emph{another} algorithm $A_2$ has expected \textsc{ots} risk below 1/2 (indeed \emph{zero}, for choice of $\mathcal{D}$ that labels instances deterministically via some $f^*$, and $A_2$ that always outputs this $f^*$).  

A variant for \textsc{iid} risk is the \textsc{NFL} theorem in the standard textbook by \citet[61ff]{ShaBen14}. Here the notion of performance is that in $\mathcal{D}$-expectation over samples of size $n$ no more than half the total number of possible instances, the algorithm's \textsc{iid} risk is smaller than $1/4$. %\footnote{Note 
  %that this success notion, in contrast to the Schaffer-Wolpert paradigm, does \emph{not} exclude already seen objects. This is the difference between error measures we indicated earlier in footnote \ref{fn:generror}, and also accounts for the value of 3/4 rather than 1/2.} 
  Correspondingly, their \textsc{NFL} theorem states that for every learning algorithm $A_1$ there is a learning situation $\mathcal{D}$ such that its expected \textsc{iid} risk $\Expec_{S \sim \mathcal{D}^n}\left[ L_{\mathcal{D}}(A_1(S)) \right]$ is higher than $1/4$, while that of another algorithm $A_2$ is lower than $1/4$ (indeed again zero). The authors write that the ``theorem states that for every learner, there exists a task on which it fails, even though that task can be successfully learned by another learner'' (ibid., 61).

Another example is given by the \textsc{NFL} theorems collected by \citet{Belxxtcs} for problems of prediction. (He calls these results ``of the \emph{absolute} variety,'' as opposed to ``measure-relative,'' which would include the original Wolpert-Schaffer results.) The learning situations in these problems are (probability measures over) \emph{infinite} sequences of binary outcomes, and he considers different types of effectively computable learning functions (namely, ``extrapolators'' that are, as in our example in sect.\ \ref{ssec:examplepred}, functions from past outcome sequences to next outcomes, and ``forecasters'' that output probabilities of next outcomes) and for both of these types various notions of good performance. In each case he derives two types of results, that are both instantiations of the general \textsc{NFL} statement that there is no universal algorithm: that for each learning algorithm $A_1$ there is a second algorithm $A_2$ that performs well in those situations in which $A_1$ does, \emph{and} in other situations still (``better-but-no-best''),\footnote{Note 
  that such results, while instances of \textsc{NFL} statement \eqref{eq:NFL}, go against the Wolpert-Schaffer statement that all algorithms are equivalent. There are in this learning framework strictly better and better performing algorithms---just no \emph{best}.} 
  and that for each $A_1$ there is an $A_2$ such that the situations in which they perform well are \emph{disjoint} (``evil-twin''). 

These examples also illustrate that statement \eqref{eq:NFL} retains much of the spirit of the original Wolpert-Schaffer statement. In particular, it is a clear expression of the central insight in machine learning  \citep{Mit80tr,Mit97,Die89icml,Rus91ps,ShaBen14} that no purely data-driven learning algorithm---no formal inductive function from data to conclusions---can be successful in all circumstances. That is, every such data-only algorithm must possess some \emph{inductive bias} that determines in which restricted class of situations it performs well, and hence in which situations it does not. What  statement \eqref{eq:NFL} still adds to this is that such a learning algorithm's inevitable inductive bias excludes it from learning successfully in some situations that are not \emph{unlearnable}: situations in which \emph{some other algorithm} would perform well. But it does not go as far as the original Wolpert-Schaffer statement that all (data-driven) algorithms are \emph{equivalent} in their performance, depending as this does on the additional and unmotivated assumption of a uniform prior distribution. 

\section{Generic algorithms and local models}\label{sec:nounivrule}
%In the previous section, we discussed the original Wolpert-Schaffer \textsc{NFL} results. We noted that these results use a conception of learning algorithms as purely data-driven, as functions from data to conclusions. Moreover, we criticized the assumption of a uniform prior distribution on learning situations, that sustains the interpretation that all algorithms are equivalent. This led us to advocate the alternative \textsc{NFL} statement \eqref{eq:NFL} that there is no universal data-only learning algorithm. 
%In this section, we investigate the significance of this statement for the justification for machine learning algorithms. 
In this section, we investigate the significance of the NFL-statement  \eqref{eq:NFL} for the justification for machine learning algorithms.  The route we take is to first relate the \textsc{NFL} results to Hume's skeptical argument about induction (sect.\ \ref{ssec:skep}). %We identify a presupposition that is central both to Hume's original argument and discussions of the original Wolpert-Schaffer results. This is the presupposition that justifying inductive methods requires justifying a general postulate of the induction-friendliness of the world.
We note that both Hume's original argument and discussions of the original Wolpert-Schaffer results presuppose that justifying inductive methods requires justifying a general postulate of the induction-friendliness of the world.
 Subsequently, we discuss philosophical work that denies this presupposition, and that promotes a \emph{local} perspective on induction (sect.\ \ref{ssec:local}). We argue that a local conception of induction, applied to machine learning, points at a more natural conception of learning algorithms: rather than one-place functions \emph{on data only}, many standard learning algorithms are better conceived of as two-place functions that for their operation also require some \emph{model} (sect.\ \ref{ssec:models}). Learning-theoretic guarantees do justify the use of such  algorithms, in a local, \emph{model-relative} manner (sect.\ \ref{ssec:modeljust}).%; even if they do not block skepticism about the absolute, global justification for the conclusions of these algorithms. 

\subsection{The road to skepticism}\label{ssec:skep}
The \textsc{NFL} theorems, both the original Wolpert-Schaffer results and instantiations of the statement \eqref{eq:NFL} of sect.\ \ref{ssec:nounivalgo}, are mathematical theorems. They say something about the impossibility of mathematically \emph{proving} that some learning algorithms, conceived of as purely data-driven, perform better %(can be expected to be generally) 
than others. As such, they can be seen as versions of the first, \emph{deductive}, horn of the fork that constitutes Hume's orginal argument against a justification for induction. This first horn concerns the impossibility of inferring good performance of inductive inference using only deductive, \emph{a priori} reasoning: since it implies no logical contradiction that induction does not perform well, %(cannot be expected to be generally) 
we can never deductively derive, from \emph{a priori} premises only, that it does.\footnote{Recall
  that ``performs well'' in \textsc{NFL} statements could mean, for instance, ``has in expectation a sufficiently high probability of a correct conclusion.'' In the analogous reconstruction of Hume's argument the deductive horn would thus amount to more than the ``boring'' observation that induction is \emph{fallible} \citep[309]{Oka01pq}: it is not just that inductive inference is not itself deductively valid and could lead to false conclusions, it does not even imply a contradiction that it is not \emph{likely} to give correct conclusions (see, e.g., \citealp[30ff]{Sky00}). The purpose of our very compressed presentation of Hume's argument here is mainly to draw analogies to the \textsc{NFL} results and their implications; and we pass over some issues regarding the proper reconstruction of Hume's original argument that are not uncontentious (including whether the argument was intended to extend to probabilistic induction, and indeed the possible differences in conceiving of the two horns as ``deductive'' v.\ ``inductive'' or as ``a priori'' v.\ ``empirical''). See \citet{Lan11incl,Hen20sep} for entries to the literature on Hume's (historical) argument. } 
  Similarly, the \textsc{NFL} results show for any learning algorithm that it implies no contradiction that this algorithm does not perform well (does not perform at least as well as other algorithms), by showing that there are \emph{a priori} possible situations in which it does not.\footnote{This 
  is clearly what the \textsc{NFL} results of type \eqref{eq:NFL} do. The original Wolpert-Schaffer results fit this statement less well, at least in the usual interpretation, because of the (non-tautological) assumption of a uniform distribution. But it fits an earlier interpretation by Wolpert, mentioned in sect.\ \ref{ssec:arequiv}: it is logically \emph{possible} that learning situations are generated by a uniform distribution, hence that no algorithm performs better than any other.}

This does not yet constitute a skeptical argument that we can offer \emph{no} rational grounds for thinking that one algorithm performs better than another. Likewise, the first horn of Hume's fork did not yet establish a skeptical conclusion about the grounds for inductive inference. Arguably, the novelty and force of Hume's argument lay in the second horn of his fork: the assertion that neither can we offer, on pain of circularity, good \emph{non}deductive or \emph{empirical} grounds  for thinking that inductive inference must perform well. Only the two horns taken together lead to the skeptical conclusion that we can offer \emph{no} rational, epistemic ground for using inductive inference: that we cannot justify induction.

%In Hume's original argument, the observation that inductive inference is not deductively valid (the ``boring'' argument that induction is \emph{fallible}, \citealp[309]{Oka01pq}) was only the first horn of his fork; the novelty and force of the argument lies in the second horn, the assertion that neither can we give a \emph{non}deductive justification. Only the two horns taken together lead to the skeptical conclusion that inductive inference is \emph{unreasonable}: that we can give \emph{no} rational reason for induction.

Perhaps the Wolpert-Schaffer results were not intended to support a skeptical conclusion, and we should read conclusions of the sort that ``methods  for  induction  to  unseen  cases  cannot  be  justified   rigorously'' \citep[264]{Sch94icml} or that ``one can not formally justify [standard learning algorithms]'' \citep[38]{Wol02inc} as merely indicating the limits of mathematically founding the performance of learning algorithms. However, something more than that is suggested in the original discussion surrounding these results, by the nods to Hume (\citealp[1341]{Wol96nc}; \citealp[264]{Sch94icml}), but also by the outlines of a move very reminiscient of Hume's. This is the idea, discussed before in sect.\ \ref{ssec:arequiv}, that the only way remaining to found the good performance of our learning algorithms is to postulate that ``the world'' (or ``nature,'' or the ``universe'') has a certain structure that guarantees this. Hume's original argument in fact \emph{starts} with the premise that inductive reasoning proceeds upon the principle that ``nature is uniform.'' It is this principle that is subjected to the two horns; in particular, that we cannot justify it \emph{inductively} or \emph{empirically}. Namely, any attempt to derive the uniformity of nature from past such observed uniformity would require the very principle at stake and thus be viciously circular.

%The first horn of his argument is that we cannot justify this principle \emph{a priori} or deductively. The second horn is that we also cannot justify the principle \emph{inductively} or \emph{empirically}. Namely, any such argument, that derives the uniformity of nature from past such observed uniformity, would require the very principle at stake and thus be viciously circular.

Hume's argument and most of its later reconstructions simply concerned ``inductive inference'' or ``inductive method,'' exemplified by something like enumerative induction but beyond that largely left unspecified (prompting a distinct problem of description, recall sect.\ \ref{ssec:allalgos}). The \textsc{NFL} theorems concern all possible purely data-driven learning algorithms. Still, the skeptical threat of the \textsc{NFL} results lies in their application to ``our standard algorithms,'' the generic learning algorithms that we actually use (recall again sect.\ \ref{ssec:allalgos}). So both Hume's argument and discussion surrounding the \textsc{NFL} results envisage some restricted collection of generic inductive methods. And in both cases we see that the performance of these inductive methods is paired to a particular structure the world may or may not have. If the world has the matching structure, then our inductive methods perform well; if not, they do not.\footnote{In 
  Hume's argument, this structure is given by the principle of the uniformity of nature; in the case of the \textsc{NFL} results, by a \emph{non}-uniform distribution. Note the opposed denotations of the term ``uniform'' here: a uniform distribution intuitively signifies complete randomness and thus lack of structure and regularities, so that it corresponds to an assumption of \emph{non}-uniformity of nature in the sense of Hume.}
  Consequently, the dialectics turns on the justification for such an assumption on the world: in Hume's argument from the start, in the case of the Wolpert-Schaffer results in the ensuing discussion. The \textsc{NFL} statement \eqref{eq:NFL} is similarly susceptible to this move: if we do want to uphold the existence of well-performing generic (\emph{universal}) learning algorithms, then it seems we must postulate that the world has a structure that facilitates such algorithms' performance. But in all cases, it appears impossible to justify, without question-begging, such an assumption on the world, whence we are driven towards a skeptical conclusion. 
  
\subsection{Localizing induction}\label{ssec:local}
An idea that has been advanced in the philosophical literature is that we may avoid being driven there by \emph{denying that inductive inference relies on universal uniformity principles} \citep{Oka05pas}. This idea builds on arguments that it is hopeless to try and give a precise account of a principle of the ``uniformity of nature'' (\citealp{Sal53ppr}; \citealp[55ff]{Sob88}).\footnote{Arguably 
  it is already the lesson of Goodman's new riddle that Hume's uniformity of nature principle is empty (\citealp[309]{Oka01pq}; \citealp[58f]{Lan11incl}; also recall footnote \ref{fn:newriddle}), % or that there are no general rules for induction (refs), 
even if this is obscured by Goodman seeking to patch Hume's supposition by a restriction to ``projectable'' predicates (\citealp{Ros82jop}; \citealp[320]{Oka01pq}).}

Sober (\citeyear[58ff]{Sob88}; also see \citealp[245ff]{Oka05pas}) argues that in presupposing that induction relies on a single principle of uniformity, Hume actually commits a \emph{quantifier-shift fallacy}.  It is not the case, as Hume has it, that there is a certain assumption (the uniformity of nature) that every inductive inference requires; it is rather the case that \emph{every inductive inference requires a certain assumption}. That is, rather than all relying on a single universal uniformity principle, every induction relies on a specific and \emph{local} empirical assumption. 

Arguments against universal uniformity principles usually run together with arguments that it is hopeless to try and give a precise account of ``inductive method'' (\citealp{Put81,Put87}; \citealp{Ros82jop}; \citealp{Fra89}; \citeyear{Fra00ppr}; \citealp{Nor03pos,Nor10pos}). Okasha \citeyearpar{Oka01pq} indeed develops an argument analogous to Sober's where  he diagnoses the fault in Hume's reasoning to be the presupposition that inductive inference is given by universally applicable rules. He, like Norton (\citeyear[666]{Nor03pos}; \citeyear{Nor14syn}), argues that the denial of this presupposition actually blocks the skeptical argument.

These ideas offer a \emph{local} perspective on inductive inference.\footnote{The
  general idea of urging a local perspective on induction can be discerned in various philosophical currents, including the pragmatist tradition (see \citealp{Lev67,Bog76}). %[The same holds for the akin idea that the relevant or at least feasible question of justification concerns the \emph{changing} of beliefs in a reasonable fashion, irrespective of the starting point.] 
  Nor is the questioning of a principle of the uniformity of nature \citep{Pei78psm,Pei02incol} or formal schemes of induction (see \citealp{Mcc21shposa}) remotely new. We focus here on the relatively recent writings by Okasha, because they specifically address the Humean problem of (global) justification, and because they bring out nicely the points that are important to our argument.} 
  In order to assess the value of this perspective for machine learning algorithms and their justification, we make two observations.

First, even if we grant that Hume's original argument no longer goes through when we deny the existence of universal uniformity principles or inductive rules, it  does not follow that we are safe from a skeptical argument. As Sober \citeyearpar[66ff]{Sob88} himself emphasises, there are still always assumptions involved in an inductive inference, that themselves stand in need of justification. Even if we are safe from Hume's argument that any nondeductive justification of induction must be \emph{circular}, it appears we will now be facing an endless \emph{regress}, where each empirical assumption can only be justified by another induction with its own empirical assumptions. 

Yet Okasha \citeyearpar{Oka05pas} is more optimistic: ``The \emph{form} which the inductive sceptic's argument takes on the $\forall\exists$ picture---pushing the demand for justification further and further back---seems somehow less problematic  than  in  the $\exists\forall$ case,'' where ``the whole practice of reasoning inductively seems to be premised  on  an  enormous,  untestable  assumption  about  the way the world is'' (ibid., 252, 251). We do not think that this settles the matter, but it does clearly bring out a crucial advantage of a local perspective on induction. Namely, this perspective is much closer to what the problem of justification looks like in \emph{actual enquiry}.\footnote{This 
  is also a main selling point of Norton's ``material theory of induction.'' But it is likewise far from clear that Norton can uphold his promise of escaping inductive skepticism, in particular, of escaping the endless justificatory regress %(\citealp[666]{Nor03pos}; \citeyear{Nor14syn}; 
  (\citealp{Kel10pos}).} 
  Plausibly, in an actual enquiry, each inference takes place within a constellation of context-specific or local empirical assumptions.\footnote{For 
  a dissenting view, based on examples from the history of science of inductions within ``theoretically impoverished contexts,'' see \citet{Lan02pq,Lan04n}.} 
  The motivation for such an inference will focus on one or more of these assumptions, and not on a universal uniformity principle. Furthermore, the question of justification does not only target these assumptions: even \emph{given} these assumptions, there can still be room for different inferences, in which case there is still the question of the justification for the inference of choice, or the method used for the inference. We will argue below that both aspects are important to the question of the justification for machine learning algorithms. %[bovendien opent dit de weg voor een rechtvaardiging van schema's, onder de aanname dat het model accuraat is---een kwestie die ook nog niet triviaal is]

Second, it might seem that a local conception of induction, inasmuch as it is coupled to the position that inductive inference cannot be encoded into general \emph{rules}, actually does not sit very well with the enterprise of machine learning. After all, and arguably in contrast to day-to-day human or even scientific reasoning, machine learning is characterized by the design and use of learning algorithms: fully mechanical, generic procedures for inductive inference.

The rejection of general inductive rules in a local perspective must be qualified, though. For instance, Okasha (\citeyear{Oka01pq}; also see \citeyear{Oka05cjp}), in the course of arguing against the idea of general rules for inductive inference, does endorse Bayesian conditionalization as the rational procedure for learning from experience.\footnote{So 
	do Rosenkrantz \citeyearpar{Ros82jop} and van Fraassen \citeyearpar{Fra89}. Norton \citeyearpar{Nor10pos,Nor14syn} does categorically argue against any formal scheme for induction (including the Bayesian scheme), which places his theory at odds with the perspective we develop here.} 
	There appears to be a tension there (cf.\ \citealp{Hen20sep}): is updating by conditionalization not a rule? Okasha, however, makes a distinction: ``a rule of inductive inference is supposed to  tell you  what  beliefs  you  should  have,  given  your  data,  and  the rule of conditionalization does not do that \dots \ the state of opinion you end up in depends on the state you were in previously; whereas if you apply an inductive rule to your data, the state of opinion you end up in depends on the instructions contained in the rule'' \citeyearpar[316]{Oka01pq}. The output of Bayesian conditionalization does not depend on the input data \emph{only}: it also depends on ``the state you were in previously,'' ultimately, a prior probability assignment. The rejection of general rules for inductive inference here thus concerns purely \emph{data-driven} rules. 
	
This idea is, of course, very much supported by the statement of the \textsc{NFL} theorems we advocated: there is no universal \emph{purely data-driven} learning algorithm.\footnote{Van 
  Fraassen's (\citeyear[132ff]{Fra89}; \citeyear[256ff]{Fra00ppr}) rejection of ``the ideal of induction'' (``a rule'' that is ``rationally compelling,'' ``objective,'' and ``ampliative'') relies for an important part on results that go back to a proof of Putnam's \citeyearpar{Put63inc1} that is in effect an instantiation of \textsc{NFL} statement~\eqref{eq:NFL}. Putnam shows by a diagonalization argument that for each prediction algorithm, there exist infinite data sequences on which this algorithm does not perform well, sequences that are in fact themselves effectively computable so that another algorithm predicts them \emph{perfectly} (also see \citealp{Ste19erk}).}
  Moreover, this is perfectly consistent with allowing for general rules for induction that also require other inputs, plausibly inputs that encode local assumptions, like (in the case of the Bayesian method) a prior probability distribution. In sum, the lesson we take from a local conception of induction is not to reject rules for induction: the lesson is to fine-grain the notion of inductive rule, to conceive of it as a procedure that can also take for input local assumptions. Applying this perspective to machine learning algorithms, we will also be able to qualify the sweeping skeptical conclusion that the \textsc{NFL} theorems seemed to lead us to.

\subsection{Model-dependent learning algorithms}\label{ssec:models}
%PETER viz gesprek met Tom hieronder heb ik sectie titel veranderd
%PETER: we hebben dus het volgende afgesproken: 
% we richten ons specifiek op de 'rare' claims 'error minimization is maximization' CV is anti-CV etc. DIE gaan over modellen of meta-modellen
% en ALS je die hebt, dan kun je beter aan een tweeplaatsige versie denken. We geven toe dat dingen als nearest neighbour en neurale netwerken
% ingewikkelder zijn en algoritme/model moeilijk te scheiden zijn - dat is future work. Maar ALS het te scheiden is, dan zijn er goede/
% slechte methoden.  
% verder nuanceert tom: in 'kleine' learning problems kunnen er soms wel universele algoritmes bestaan. Daar moet iets over in in intro?
% EERST Bayes - statistical model, dan LEARNING THEORY "learn the best within a set"
We think it highly implausible that the use of machine learning algorithms relies, explicitly or implicitly, on a general ``assumption of machine learning''  %The use of standard learning algorithms is not motivated by a postulate 
about the learning-friendliness of the world, let alone a belief in some all-governing non-uniform prior distribution on possible learning situations. The assumptions that accompany the use of a learning algorithm in any particular context are normally themselves of a context-dependent, \emph{local} nature. %The use of a learning algorithm in any particular context rather relies on particular assumptions, \emph{local} assumptions that one makes within that particular context. 
But how to square the role of local assumptions with the use of generic  mechanical learning procedures?
%PETER to TOM: waarom staat er een , achter generic?
%TOM to PETER: leest misschien beter? maar verder geen noodzaak

The observations of the previous section point us at an answer. Many standard learning algorithms are not purely data-driven, but must also take for input a \emph{model}. Such \emph{model-dependent} algorithms instantiate, not a one-place function that maps data to conclusions, but a \emph{two}-place function that maps data \emph{and a model} to conclusions. Crucially, such algorithms can be given a \emph{model-relative} justification.

In the following, we illustrate model-dependent learning algorithms using three standard machine learning examples: Bayesian machine learning (sect.~\ref{sssec:modelbay}), empirical risk minimization (sect.~\ref{sssec:modelerm}) and cross-validation (sect.~\ref{sssec:modelcv}). 
%PTEER TO TOM 17-5: added following sentence: 
These methods all have in common (as do most if not all standard model-dependent learning algorithms that we know of) that they select a hypothesis or combination of hypotheses with good predictive performance, measured in terms of the loss function of interest (empirical risk minimization, cross-validation) or a related measure such as the likelihood (Bayes).
We discuss how these methods receive a model-relative justification in the form of learning-theoretic guarantees, and thereby bring out why such claims as ``the \textsc{NFL} theorems indicate that cross-validation has no more inherent justification than anti-cross-validation'' are misleading.\footnote{We
	here say very little about potentially useful distinctions between different \emph{types} of justification and the exact nature of the accompanying inductive assumptions, but this is a natural avenue for further investigation \citep{Cor10mm}.    
  } We conclude our examples with a discussion of the consistency of the \textsc{NFL} results with learning guarantees (section \ref{sssec:modelnfl}). %and the model-relative justification of learning algorithms (section \ref{sssec:modeljust}). 

Finally, we have delegated to appendix~\ref{app:details} some nuances %and technical details 
that distract from the argument's main thrust.% (specifically, the limitations  of standard model-dependent algorithms, and the existence of standard algorithms for which the model-dependent perspective is more tenuous).

%address potential loose ends in our agument, 

%TOM to PETER: ik heb me in de revisie van jouw tekst vooral beperkt tot de notatie/begrippen consistent maken met het voorgaande en sommige dingen wat bondiger formuleren. controleer vooral even of het nog steeds allemaal klopt!
%plaatsen met (voorgestelde) grotere wijzigingen zijn voorzien van aparte comments
%PETER2 ik heb vooral bij CV stukje toch weer wat dingen veranderd in de wiskundige formules - nu moet jij nog even checken of het allemaal logisch is wat er staat!

\subsubsection{Bayesian learning}\label{sssec:modelbay}
The Bayesian scheme, central to many philosophical accounts of rational learning, also constitutes an important approach in machine learning
\citep{DudHarSto01,bishop2006pattern}. 
%PETER to TOM: bovenstaande zin misschien overkill en weghalen?
%TOM to PETER: wat bondiger geformuleerd. De waarde ervan is een (subtiele) herinnering aan de vorige sectie dat het lokale perspectief op Bayesiaans redeneren in de filosofische context noemde.
What characterizes Bayesian learning is that an algorithm must be provided with a prior distribution over some domain of probability distributions, and this choice of prior constitutes a choice of model. The role of the prior as a variable input factor lends such an algorithm a considerable genericity: the algorithm \emph{itself} does not come with a particular model, but is flexibly supplied with a specific model in each specific application. %The algorithm \emph{itself} can lay claim to genericity, because via the choice of model it can be made to run with a wide variety of local assumptions. 
This is also what provides room for a \emph{model-relative} learning-theoretic justification: a demonstration that \emph{relative to} the choice of prior distribution, a Bayesian algorithm performs well. 

We now discuss this in some detail for Bayesian machine learning in the framework of classification,  the realm of the original Wolpert-Schaffer results. Here, the prior $\Pi$ is usually taken over a set of {\em conditional\/} probability distributions  of the form $P(Y\mid X)$ with $Y \in \mathcal{Y}$, $X \in \mathcal{X}$ the possible labels and instances, respectively. (Recall the oatmeal example of sect.~\ref{ssec:exampleclassif}, where $\mathcal{Y} = \{ \mathsf{T,N}\}$ and $\mathcal{X} = \{0,1\}^3$.) 
The distributions are extended to $n$ outcomes by assuming that the data pairs  $(X_1,Y_1), (X_2,Y_2), \ldots, (X_n,Y_n)$ in the training set are sampled independently. The set of probability distributions in the prior support (that is, those with prior density or mass greater than $ 0$) demarcate a model $\mathcal{M}$, a set of (conditional) probability distributions. A prototypical example is the logistic regression model \citep[119ff]{HasTibFri09}, in which the $X_i = (X_{i,(1)}, \ldots, X_{i,(k)})$ are vector-valued as in our example, and 
the $P(Y \mid X)$ are given by linear functions $\sum_{j=1}^k \beta_{(j)} X_{(j)}$, rescaled by a fixed nonlinear function so as to become probabilities that sum to one. 

There exist several variations of the Bayesian stance, which differ in how the prior is interpreted. For the purpose of our discussion, most relevant is the distinction between a \emph{subjective} and a \emph{pragmatic} stance. Under the former, the prior quite literally encodes one's beliefs (which can be elicited by, for example, testing willingness to bet on certain outcomes). That is, the relevant inductive assumption can be equated with one's beliefs. Alternatively, under a pragmatic interpretation, to which most practitioners subscribe, one still assumes the {\em model\/} (set of all distributions in the support of the prior) to be correct, but one can choose the prior $\Pi$ for other, more pragmatic reasons. These could be considerations of (computational) convenience, of optimizing worst-case behaviour (this leads to ``noninformative'' or ``flat'' priors), or a mix of prior knowledge with worst-case and computational considerations. For example, a standard pragmatic approach for the logistic regression model is to take a Gaussian prior centered at $0$ on the $\beta_{(j)}$'s. 

 %reasons of (computational) convenience, or for optimizing worst-case behaviour so that  $\hat{f}_{\text{\sc Bayes}}$ has the smallest possible generalization error in the worst case under all $P^* \in \mathcal{M}$ (this leads to `noninformative' or `flat' priors), or by a mix of prior knowledge with worst-case and computational considerations. 

%Under the subjective interpretation of Bayesian inference, the prior $\Pi$ that encodes one's inductive assumptions precisely describes one's beliefs. Alternatively, under the pragmatic view of Bayesian learning to which most practitioners subscribe, one may view one's inductive assumption simply as the model $\mathcal{M}$, 
%TOM to PETER: dit is een beetje onduidelijk: het klinkt alsof je zegt dat we wél geloven in de waarheid van het model, maar niet zozeer in de precieze piorverdeling daarover?
%PETER2: dat is inderdaad precies wat ik bedoel. We zouden kunnen toevoegen (iets als):
%'--- we do believe the model is correct, but we do not interpet the prior $\Pi$ as indicating true beliefs about the relative probability of elements of $\mathcal{M}$ being correct.' Ik laat het aan jou over om te beslissen of we dat echt moeten doen. 

Regardless of the prior's origin, it serves as an input to the Bayesian algorithm. Together with the data, i.e., the training sample, 
one uses Bayes's rule to update the prior to a posterior. 
The posterior over the distributions is then used to output a classifier $\hat{f}_{\text{\sc Bayes}}$, defined as the function from $\mathcal{X}$ to $\mathcal{Y}$ that  has the largest probability of being correct according to the Bayesian posterior predictive distribution \citep{bishop2006pattern}. 
%In the degenerate case where all distributions in ${\cal M}$ are really `deterministic', assigning, for each $x$, probability $1$ to one of %the two choices for $Y$, this amounts to a weighted majority vote. 
In contrast to the notion of learning algorithm in sect.\ \ref{sec:equiv}, where an algorithm only takes data, the Bayesian algorithm requires additional input: 
the user's inductive assumptions, codified explicitly as prior and induced model. One cannot avoid stating these explicitly---without specifying a prior and hence a model, the outcome of the Bayesian algorithm is simply undefined.

When it comes to the question of justification, the distinction between the two Bayesian stances is also
%PETER TO TOM 17-5 ADDED FOOTNOTE KUN JIJ NOG REFERENTIE NAAR BIJV DE FINETTI TOEVOEGEN?
 relevant. %\footnote{As a referee remarked, for ``hard-core'' (to use the referee's terminology) subjective Bayesians the two justifications can be unified: such Bayesians simply insist on {\em coherent\/} decision-making. Results such as de Finetti's Dutch-book theorems essentially state that the only coherent learning algorithms are Bayesian ones; two coherent learning algorithms can only differ in terms of the choice of prior. Thus, from this point of view, {\em optimal\/} decision-making is not even an issue:  once a prior is given, the only coherent way to combine it with the data to arrive at a prediction or classification is to follow the Bayesian algorithm. }  
 Under the subjective stance, the Bayesian algorithm is simply  {\em optimal\/}: among all algorithms,  it leads to the best possible classifier (with smallest risk) under one's own inductive assumptions as encoded by the prior. In other words, if the prior truly reflects one's beliefs, then one must also believe that the Bayesian procedure, with this prior, is justified. \emph{If} one is willing to take the subjective stance, then any arguing that the Bayesian algorithm has no more inherent justification than any other algorithm, let alone ``anti-Bayesian learning'' (where one selects the classifier with the \emph{highest} risk under the posterior), is futile.\footnote{As a referee remarked, for a ``hard-core'' (to use the referee's terminology) subjective Bayesian there is a much more fundamental justification than optimality: that only Bayesian learning is \emph{rational} (made precise in terms of \emph{coherence}, or quantifying uncertainty in terms of degrees of belief that satisfy the probability axioms). For instance, \citet[21]{bishop2006pattern} writes, ``The use of probability to represent uncertainty \dots\ is inevitable if we are to respect common sense while making rational coherent inferences.'' From this point of view, performing well is strictly speaking not even an issue:  once a prior is formulated, the only coherent and therefore reasonable way to combine it with the data to arrive at a prediction or classification is to follow the Bayesian algorithm. }

Under a pragmatic view of Bayesian inference, the prior weights cannot be directly related to one's beliefs, and the Bayesian algorithm cannot be said to be optimal in the previous sense. 
%PETER2 to TOM: Tom schreef over 1e versie 'maar hoe zit het dan met al die resultaten dat een Bayesiaan over-optimistisch is en we niet teveel waarde moeten toekennen aan zijn prior-inschattingen van zijn eigen prestaties? [dit is ook een steeds groter thema voor filosofen]
%Peter antwoordt: ik denk dat dat nu ondervangen is door direct te zeggen 'not necessarily optimal'. Maar als je wilt kunnen we er ook directe citaten naar over-optimistisch gedrag e.d. inzetten (dat zouden dan misschien wel papers moeten zijn waar filosofen al wat over gezegd hebben)
Nevertheless, under the pragmatic view one can still show that the Bayesian procedure has a certain model-relative optimality, even if the specific choice of prior over the same model now becomes important. We already mentioned how choices of noninformative priors can optimize worst-case behavior, by which we meant that   $\hat{f}_{\text{\sc Bayes}}$ has the smallest possible generalization error in the worst case under all $P^* \in \mathcal{M}$.
 %reasons of (computational) convenience, or for optimizing worst-case behaviour so that  $\hat{f}_{\text{\sc Bayes}}$ has the smallest possible generalization error in the worst case under all $P^* \in \mathcal{M}$ (this leads to `noninformative' or `flat' priors), or by a mix of prior knowledge with worst-case and computational considerations. 
%Even under this pragmatic view, one can still argue that Bayesian inference has much more justification than anti-Bayesian inference. 
Furthermore, there exists a plethora of results (e.g., \citealp{GhosalGV00,GhosalLV08}) showing that, under very weak conditions on the model $\mathcal{M}$, one can select priors such that  for all $P^* \in \mathcal{M}$, the posterior concentrates around $P^*$ at a certain rate. In our context, this implies that  the expected generalization error of $\hat{f}_{\text{\sc Bayes}}$ converges to the generalization error one could obtain if one knew the ``true'' (leading to the best possible predictions) $P^*$. Moreover, one can give nonasymptotic bounds on the difference in generalization errors \citep{grunwald2016fast}. These results provide a clear model-relative justification for the pragmatic Bayesian procedure: \emph{if} one has reason to believe that the model is correct, then (with 
%PETER3 the right choice of prior
the right choice of prior over this model) %such as those advocated by the aforementioned references) 
one also has reason to believe that the algorithm performs well. 

 %which one cannot give for anti-Bayesian learning. 
%PETER2 expres 'quite similar' weggehaald (omdat het verschil van 'model als kansverdeling vs model als predictor' toch nogal groot is):
%
%
%The story is quite similar to

For the sake of brevity we do not go in more detail into the justification of Bayesian methods. Instead, we proceed with a more in-depth discussion of two methods that have received more attention in the context of the \textsc{NFL} results: empirical risk minimization and cross-validation.
%

%TOM to PETER: ik zat te denken dat er grofweg twee manieren zijn waarop je een "model-relatieve" rechtvaardiging op kan vatten:
%--als je model waar/accuraat is, dan werkt je methode goed (maar als het niet accuraat is, kan je methode heel slecht zijn---zelfs een stuk slechter dan het beste in je model!)
%--je methode werkt zo goed als het beste in je model (als je model slecht is, doet je methode dat ook---maar niet slechter)
%dit lijken relevante dimensies voor de discussie over model-relatieve rechtvaardiging in Bayes en in leertheorie, misschien dit expliciet maken? 
%PETER2: ja dit klopt helemaal. Maar de vraag is **waar** we dit moeten opschrijven, daar kom ik niet uit. Op deze plek in de tekst (net voor ERM) kan het nog niet. Ik heb het nu dus nog maar niet gedaan en 'relevante' stukje hieronder staat dus nog in appendix. Een optie is de volgende: we vervangen de laatste zin
%
%The story is quite similar to the one for \textsc{ERM} and cross-validation, so we omit the details and now turn to these. 
%
%door
%
%The story is analogous to the one for ERM, which is however more easily explained, so we now turn to ERM first and then retun to 
%pragmatic Bayes in Section 3.3.3. . Dan nieuwe Sectie 3.3.3. (na ERM, voor CV) met als titel Pragmatic Bayes vs ERM. Daar hebben we het dan %over die 2 vormen van 'model-relatieve rechtvaardiging', en noemen vast even dat als je probabilistische model incorrect is, dingen verkeerd
%kunnen gaan met Bayes. 
%
%maar zo'n extra subsectie is misschien beetje rommelig?

\subsubsection{Empirical risk minimization}\label{sssec:modelerm}
This is probably the most standard machine learning method. Like Bayesian learning, empirical risk minimization (\textsc{ERM}) is a model-dependent method. The crucial difference with Bayesian learning is that the ``model'' is now not a set of probability distributions, but rather a user-specified set of classifiers $\mathcal{F}$, usually called a {\em hypothesis class}. %This class is user-specified and can be any set of classifiers. 
In practice, it could be the set of all neural networks with a given number of nodes and connectivity matrix, represented by their weights; or the set of all decision trees of a given size.
The generalization performance of \textsc{ERM} 
can be analyzed via the standard machinery of learning theory \citep{ShaBen14}. Here, as in sect.\ \ref{ssec:exampleclassif}, one assumes that the data $S = (X_1, Y_1), (X_2, Y_2), \ldots, (X_n,Y_n)$ are sampled independently from an unknown distribution $\mathcal{D}$. 
No further assumptions about $\mathcal{D}$ are made: instead all inductive assumptions go into  $\mathcal{F}$. In Bayesian learning, the choice of model $\mathcal{M}$ can be seen as the inductive assumption that ``there is a $P \in \mathcal{M}$ such that acting as if the data is a random sample from $P$ leads to the best possible predictions.'' In learning theory, adopting a class $\mathcal{F}$ can be seen as the inductive assumption that ``there is an $f \in \mathcal{F}$ that has classification risk $\ll 1/2$, small enough to be useful.'' Here the classification risk is \textsc{iid} risk, or the probability that $f(X) \neq Y$ under $\mathcal{D}$.
%PETER to TOM: probably need to rename the term 'classification error' and/or this explanation so as to be more aligned with the rest of the manuscript  . Also this may need to go to more central place
%TOM to PETER: ik definieer en leg "iid risk" nu uit in sectie 2.2

The \textsc{ERM} method $A_{\textsc{erm}}$ takes as input 
%PETER2 we gebruiken nu hier A_erm(S,\mathcal F} en in de appendix
% ERM_mathcal{F}(S). Is dat een probleem? misschien beter
% om consistent te zijn (ik vind allebei ok) 
%TOM2: ja goeie, ik heb het in de appendix aangepast.
both a training sample $S$ and a hypothesis class $\mathcal{F}$ as above. 
It proceeds by picking the classifier $\hat{f}_{\text{\sc erm}} = A_ \textsc{erm}(S,\mathcal{F})$ in  $\mathcal{F}$ that made, among all elements in $\mathcal{F}$, the minimum number of errors on $S$, with some arbitrary rule for breaking ties. Assume for simplicity that $\mathcal{F}$ is finite, so that there exists an $f^*$ in $\mathcal{F}$ that minimizes the risk. A variation of a standard result in learning theory says that \textsc{ERM} works well, in the following sense: the difference between the expected risk of $\hat{f}_{\text{\sc erm}}$ and the best obtainable risk within the model, namely that of $f^*$, is bounded by $\sqrt{| \log \mathcal{F}|/(2n)}$. (See appendix~\ref{ssec:derivation} for a derivation.) 
%PETER to TOM I still need to check the constant 8. It is either 4 or 8...
%..and I need to provide a reference. 
%PETER2: de constante is fijner dan ik dacht. Ik doe de afleiding nu 'even snel' in de appendix 'for completeness' omdat ik letterlijk bewijs niet kon vinden in de literatuur, en omdat dit ook nodig is voor het result voor CV (of eigenlijk: 2-forward validation)
This result holds no matter what $\mathcal{D}$ is. 
Since the dependence on the size of $\mathcal{F}$ is logarithmic, the guarantee remains non-void even for exponentially large, and in this sense fairly complex $\mathcal{F}$. In fact, it can be extended to many infinite $\mathcal{F}$ as well: the $\log |\mathcal{F} |$ term is then replaced in the bound by an abstract (but computable) complexity notion such as the Rademacher, Vapnik-Chervonenkis or ``PAC-Bayesian'' complexity of $\mathcal{F}$ \citep{grunwald2016fast}. Interestingly, as the latter paper explains in detail, such results are proven using essentially the same techniques as those used for proving non-asymptotic convergence of pragmatic Bayesian learning.

What about anti-\textsc{ERM} (or empirical risk \emph{maximization}), that picks the $\hat{f}_{\text{\sc a-ERM}} \in \mathcal{F}$ with \emph{largest} error on the training set? We can precisely reverse the math behind the convergence of \textsc{ERM} to show that anti-\textsc{ERM} will converge to the {\em worst\/} element of $\mathcal{F}$, the element that maximizes risk.
The difference between the expected risk of $\hat{f}_{\text{\sc a-ERM}}$ and the worst obtainable risk is again at most $\sqrt{| \log \mathcal{F}|/(2n) }$ if $\mathcal{F}$ is finite, and an analogous result holds again with $\log |\mathcal{F} |$ replaced by Rademacher or VC dimension for infinite $\mathcal{F}$. 
Saying ``\textsc{ERM} has no inherently better justification
than anti-\textsc{ERM}'' would thus amount to saying:  ``A method which (given a not too small sample) leads to the best possible predictions that can be obtained based on my hypothesis class, has no more inherent justification than a method which (given a not too small sample) leads to the worst possible predictions that can be obtained based on my hypothesis class.'' To us, this seems an aberration.%PETER2 nieuwe voetnote leek hier wel op zn plaats maar kan misschien strakker geformuleerd
\footnote{A complication is that the original \textsc{NFL} claim ``\textsc{ERM} is no more justified than anti-\textsc{ERM}'' is based on measuring classification quality in terms of \textsc{ots} error, whereas the learning-theoretic claims that \textsc{ERM} is better are based on \textsc{iid} error. In appendix~\ref{ssec:appwolots} we lift this complication by explaining that in practice, both error measures often essentially coincide.} 

Our point is certainly not that \textsc{ERM} is perfect: if $\mathcal{F}$ becomes ``too complex'' then \textsc{ERM}  may suffer from severe overfitting and will not work in practice.\footnote{See appendix~\ref{ssec:appnuanlims} for more details about shortcomings of \textsc{ERM} as well as cross-validation.} But if anyone advises us to use such a class in combination with \textsc{ERM}, we can simply reply that handling it goes beyond the power of \textsc{ERM}---other methods more suitable for that case exist, such as structural risk minimization \citep{Vap98,ShaBen14}, or forms of minimum description length learning \citep{GrunwaldR20}, or \textsc{ERM} combined with cross-validation as below.

We thus have a well-defined condition (small enough complexity of our $\mathcal{F}$) under which \textsc{ERM} is provably preferable to anti-\textsc{ERM}. No such conditions have ever been formulated under which anti-\textsc{ERM} performs better than \textsc{ERM} (with the same model!), and it is highly implausible that something of the sort could be done.

\subsubsection{Cross-validation}\label{sssec:modelcv}
This method can be viewed as a meta-algorithm to select {\em between\/} different learning algorithms.\footnote{Cross-validation is often seen as a highly non-Bayesian method, but there are in fact close connections, as first pointed out by \citet{Daw84jrss}; also see \citet{FongHolmes2020crossvalidation}.}
For ease of presentation, we concentrate on a simplification of cross-validation: {\em two-fold forward-validation}. This takes as input a data set of a given size $n>1$, and a finite 
%(but possibly dependent on $n$)
set of learning algorithms $A_1, A_2, \ldots, A_{m}$.  Forward-validation runs all these algorithms on the first half $S_1$ of the original training set.\footnote{If $n$ is odd, we take $S_1$ to contain the first $(n-1)/2$ data points and $S_2$ the remaining ones.}   Letting $\hat{f}_{k}=A_k(S_1)$ denote the classifier learned by algorithm $A_k$, it then selects as final classifier the classifier 
$\hat{f}_{\hat{k}_{\text{fv}}}$ 
%
%$=A_{\textsc{cv}}(S_2,(A_1,\dots,A_k))$, 
%PETER2 to TOM bovenstaande klopt niet helemaal... het zou moeten zijn:
%
%$\hat{f}_{\hat{k}_{\text{fv}}}=A_{\textsc{fv}}(S_2,(A_1(S_1),\dots,A_{m}(S_1)))$, 
%
%maar ik weet niet of we A_{cv} uberhaupt moeten introduceren dus heb het voor nu weggelaten. Als je denkt dat het helderder is,
%prima (zie ook hieronder), maar dan wel op de manier zoals hier staat, met expliciete vermelding van S1...)
%TOM2 waarschijnlijk inderdaad beter zonder
where $\hat{k}_{\text{fv}}$ is the $k$ such that 
$\hat{f}_k$ has the smallest error on the second half $S_2$ of the training set, which is thereby used as a \emph{validation} set. 
%PETER added 14-15 in response to referee 2
Thus, the final classifier always coincides with one of the $m$ initial classifiers.
For full two-fold cross-validation, one repeats the procedure with the two data sets interchanged, and for $M$-fold cross-validation we split the data in $M$ parts with a validation set of size $n/M$. Everything we say below for two-fold forward-validation also holds {\em mutatis mutandis\/} for full $M$-fold cross-validation, but the phrasing of results becomes more cumbersome, so we stick to the two-fold forward case for simplicity. 
%\commentout{ 
%For full two-fold cross-validation, one repeats the procedure with the initial training and validation set interchanged. This results in an additional set of classifiers $\hat{f}'_{k}$ denoting the classifiers learned by algorithm $A_k$ using the second half of the data as training set. One then chooses $\hat{k}_{\text{cv}}$ to be the $k$ such that the sum of the errors that $\hat{f}_k$ makes on the second half of the training set and $\hat{f}'_k$ makes on the first set is minimized. 
%The final classifier $\hat{f}_{\text{cv}}$ that is output is the {\em probabilistic\/} classifier which works by first tossing a fair coin; if the coin lands heads, it classifies $x$ as $\hat{f}_{\hat{k}_{\text{cv}}}(x)$; if tails, it classifies $x$ as $\hat{f}'_{\hat{k}_{\text{cv}}}(x)$.
%}

%PETER2: gezien je opmerkingen hieronder was het niet helemaal helder wat ik schreef. Ik heb nu alles herschreven
%gebruikmakend van jouw explicietere definities, check even goed of het nu duidelijk is.
%TOM2: ja, helder! 
Now, let $\mathcal{E}^{(n)}_k$ be the expected \textsc{iid} risk of algorithm $k$ after having run on the first half of the data: $\mathcal{E}^{(n)}_k =
\Expec_{S \sim \mathcal{D}^n}\left[L_\mathcal{D}\left( \hat{f}_k \right)\right]$. 
%\Expec_{S \sim \mathcal{D}^n}\left[L_\mathcal{D}\left(A_k(S_1) \right)\right]$. 
Let $\mathcal{E}^{(n)}_{\text{fv}}$ be the expected \textsc{iid} risk 
of two-fold forward-validation as defined above:
$\mathcal{E}^{(n)}_{\text{fv}} = \Expec_{S \sim \mathcal{D}^n}\left[L_\mathcal{D}\left(\hat{f}_{\hat{k}_{\text{fv}}}\right)\right]$.
%of $\hat{f}_{\hat{k}_{\text{\sc fv}}}$ 
%PETER2: if you think it is clearer we can use $A_{\text{\sc fv}}(S_2,A_1(S_1), \ldots, A_m(S_1))$ here instead of $\hat{f}_...$, and then also use A_fv above
One can now show (see appendix~\ref{ssec:derivation}) that the expected \textsc{iid} risk of forward-validation satisfies
\begin{equation}
\mathcal{E}^{(n)}_{\text{fv}} \leq 
\min_{k \in \{1, \ldots, m \}} \mathcal{E}^{(n)}_k 
%+ \sqrt{\frac{\log m}{n}} \leq 
 + \sqrt{\frac{\log m}{n}}.
\end{equation}
%TOM2: de term aan de linkerkant was \mathcal{E}^{(n)}_k, veranderd naar \mathcal{E}^{(n)}_{\text{fv}}
%TOM2: eerst term aan de rechterkant moet niet subscript k maar \hat{k}_{\text{fv}} hebben?

%TOM to PETER: merk ook op dat ik de verwachtingen hier en elders meer expliciet opschrijf, omdat dat anders ook verwarring kan veroorzaken [er worden door het hele verhaal minstens drie soorten verwachtingen gebruikt: Wolperts uniforme verwachting over targets, de verwachting over traingsverzamelingen, en de verwachting "binnen" de risk = D-verwachte error]
%maar vooral in deze sectie ben ik niet helemaal zeker dat ik het allemaal juist "vertaald" heb, dus kun je hier even goed naar kijken? zie i.h.b. het volgende comment
Thus, the expected \textsc{iid} risk of forward-validation converges, as $n$ grows, to the expected risk of the learning algorithm that, among all algorithms under consideration, is best in the sense that it outputs the lowest-risk classifier in expectation over the training set $S_1$. 
%TOM2: complete training set $S$?
%We briefly show this in appendix~\ref{ssec:derivation}.
%TOM to PETER: in de meer expliciete notatie die ik nu gebruik, moet deze term zijn:
%$\Expec_{S \sim \mathcal{D}^n}\left[\min_{k \in m_n} L_\mathcal{D}\left(\hat{f}_{\hat{f}_{k}}\right)\right]$
%? die komt het best overeen met je omschrijving, maar niet met je formule...
%PETER2 bovenstaande commentaar van jou begrijp ik niet helemaal, maar ik denk dat het gewoon niet helder was wat ik zelf had opgeschreven
%The difference $\mathcal{E}(\hat{f}_{\hat{k}_{\text{\sc fv}}})- 
%\min_{k \in m} \mathcal{E}(\hat{f}_{k})$ is bounded by, as 
 This holds for all $m$ and $n$, so if $n$ is large, we can also take $m$ very large; in particular, due to the logarithmic dependence on $m$, at sample size $n$ we can choose between a number of learning algorithms $m$ that is orders of magnitudes larger than $n$ and still have a meaningful bound.\footnote{For $M$-fold cross-validation, the constant in front of the square root changes from $1$ to another positive value, but remains easily computable as long as $M$ does not depend on $n$. For leave-one-out cross validation, $M=n-1$ and the mathematical analysis is tricky and a subject of ongoing research,
so we will stick here to the fixed $M$ case.}
%we can take $m$ much larger than $n$.,   for example, $m= n^4$  we see that  `converges' means that Thus, as long as the set of considered %learning algorithms $m_n$ grows subexponentially in $n$, we get a meaningful generalization bound that converges to $0$ as $n$ increases at %an explicitly computable rate.
%TOM to PETER: de uitspraak over groeiende m_n is wat verwarrend: het klinkt alsof we ons voorstellen dat we dynamisch nieuwe algoritmes toevoegen naarmate we meer data krijgen. misschien herformuleren/uitleggen? 
%PETER2: is gerepareerd

Forward- and cross-validation can be fruitfully applied both to model-dependent algorithms and to algorithms that may be better viewed as data-only.
A prototypical example of the latter is nearest-neighbor classification. Here $\mathcal{X}$ is a space equipped with a metric (e.g., Euclidean space with the Euclidean metric). The $k$-nearest-neighbor method based on a training set with $n'$ instances plus labels $(x_1,y_1), \ldots, (x_{n'}, y_{n'})$ outputs the classifier which, for any value of $x$, picks the $k$ data points $\{i_1, i_2, \ldots, i_{k} \} \subset \{1, \ldots, n'\}$ for which $x_i$ is closest to $x$, and outputs the majority vote for the corresponding $y_{i_1}, \ldots, y_{i_k}$. Nearest-neighbor with $k=1$ always has zero error on the training set, so typically overfits dramatically. However, one can use cross- or forward-validation to choose a value of $k$. The number $m_n$ of $k$'s that make sense at sample size $n$ is at most $n$, so the generalization bounds above are meaningful, and we have the guarantee that the expected risk based on using $\hat{k}_{\text{fv}}$-nearest-neighbour is close to the error achieved with the unknown optimal $k \in m_n$ that achieves the best expected risk $\Expec_{S \sim \mathcal{D}^n}\left[L_\mathcal{D}(\hat{f}_{k})\right]$. 

When applying forward- and cross-validation to model-dependent learning algorithms, one typically takes the same learning algorithm (say \textsc{ERM}) for $A_1, \ldots, A_{m}$, turned into one-place algorithms by combining each $A_k$ with a different hypothesis class $\mathcal{F}_k$. For example, $A_k$ could represent \textsc{ERM} applied to $\mathcal{F}_k$, the set of decision trees of depth $k$. The class of all decision trees of arbitrary depth is too large for \textsc{ERM} to work well (yield nontrivial generalization guarantees), but in combination with forward- or cross-validation one can use the above result to get meaningful generalization guarantees again. 

How about anti-cross-validation? We can invoke precisely the same analysis as for \textsc{ERM}. Our inductive bias is now explicitly specified at a meta-level, by specifying the algorithms $A_k$. If $m$, the number of algorithms taken into account, is fixed or grows subexponentially with $n$, cross-validation can be expected to converge to the best of them based on a finite and quantifiable sample size. In contrast, under the same conditions,  anti-cross-validation will converge to the {\em worst\/} of them. Analogously to the \textsc{ERM} case, there is a clear condition ($m$ subexponential as function of $n$) under which cross-validation is (much) better than anti-cross-validation relative to the given algorithms that encode our inductive bias. And again, we cannot imagine a condition that would allow one to prove an interesting guarantee in support of anti-cross-validation.

\subsubsection{The consistency with no-free-lunch}\label{sssec:modelnfl}
To conclude our examples, we note that the model-dependent perspective still encapsulates the valid lesson from \textsc{NFL} results: the lesson that every algorithm, when operating \emph{on data only}, must incorporate an inductive bias. A Bayesian algorithm, \emph{when provided with a model and a prior distribution on this model}, will possess a certain bias; similarly, \textsc{ERM}, {\em when provided with a hypothesis class}, and cross-validation, {\em when provided with a set of hypothesis classes}, possess a certain bias. The models here represent an inductive bias, and \textsc{NFL} results show that any such model must indeed be biased in the sense that it must be restrictive. Any algorithm plus instantiated model performs well in some situations: those situations which the inductive bias, in some sense, is well-aligned with. But the algorithm plus this model does not perform well in other situations, situations even in which the very \emph{same} algorithm, with a \emph{different} instantiated model, would perform well. %The model-dependent perspective makes clear that the algorithm itself possesses a genericity insofar as it is not itself biased, prior to being provided with a particular model.

To further illustrate the consistency of negative \textsc{NFL} results and positive learning-theoretic results, recall the \textsc{NFL} version of \citet{ShaBen14} that we described in sect.\ \ref{ssec:nounivalgo}. It states that every data-only algorithm (like \textsc{ERM} with any instantiated $\mathcal{F}$) does not perform well in situations $\mathcal{D}$ in which another data-only algorithm does perform well. They prove this by exhibiting a second algorithm that has an $\mathcal{D}$-expected risk at least 1/4 less than the first algorithm; specifically, the second algorithm is \textsc{ERM} with a class $\mathcal{F}'$ that is well (indeed perfectly) aligned with $\mathcal{D}$. Note that if the first algorithm is $\textsc{ERM}$ with some $\mathcal{F}$, then this second $\mathcal{F}'$ \emph{must} be a different class, for any significant difference in expected risk (depending on the sample size). This follows from the learning-theoretic guarantee  that the expected risk of \textsc{ERM} cannot be much worse than that of the \emph{best} hypothesis in $\mathcal{F}$, and therefore than that of any algorithm that uses (must select a classifier from) model $\mathcal{F}$. Again, \textsc{ERM} with a particular $\mathcal{F}$ may be much worse than a different data-only algorithm if $\mathcal{F}$ is not a good model. But \textsc{ERM} cannot perform much worse than any algorithm \emph{with the same model}; and \emph{if} we have reason to believe that our model is good, then we have reason to believe that \textsc{ERM} with this model performs well, too.\footnote{Our discussion here does not yet fully resolve the consistency of the \emph{original} \textsc{NFL} results with learning theory. Namely, according to these results, under a uniform prior over learning situations, any two data-only algorithms (including \textsc{ERM} and anti-\textsc{ERM} \emph{with the same hypothesis class}) are equally good. In fact, the results also imply a variant of \eqref{eq:NFL} where we drop the problematic uniformity assumption, namely that for any two algorithms (again, including \textsc{ERM} and anti-\textsc{ERM} with the same $\mathcal{F}$) there exists a $\mathcal{D}$ where the second does better than the first. In appendix~\ref{app:orignfl} we explain how these results can be consistent with the positive guarantees from learning theory.}
%PETER2 heel goed, stukje hierboven!

%TOM2: de volgende sectie buiten (de erg lange) 3.3 gehaald en aparte 3.4 gemaakt.
\subsection{The justification for learning algorithms}\label{ssec:modeljust}
Learning theory thus provides us with \emph{model-relative} justification for many standard methods. For a generic model-dependent method, %Insofar a method is generic and not itself inherently biased,
such a model-relative justification is all we can ask for. For  such a generic method, %a generic model-dependent method,
it simply does not make sense to speculate about empirical assumptions that would render the method \emph{in itself} successful and in that sense justify it. This observation stands in sharp contrast to the reduction of the justification for standard learning methods to some postulate about the right structure of the world. We think that this observation within the domain of machine learning also lends further plausibility to local accounts of induction in philosophy.

One could object, however, that no method is \emph{perfectly} generic, and some assumptions or biases are always inherent to it. To put this point differently, we have used the word ``inductive bias'' in a relatively narrow way, as only pertaining to the choice of hypothesis class. But one could object that, for instance, the method of \textsc{ERM} (anti-\textsc{ERM}), irrespective of the hypothesis class, embodies a substantive assumption that the evidence so far is not (is) misleading.\footnote{We 
thank one of the reviewers for pressing us on this point.} 
	We agree these can also be called biases, or perhaps rather meta-biases (as they concern extrapolating classifiers' success rather than the data directly); but they are fundamentally linked to assumptions that are already introduced in the formulation of the relevant learning problem, in this case the general problem of stochastic classification.\footnote{Casting
    a particular real-world problem as a particular formal learning problem (which includes, e.g., choice of possible instances and labels) is, of course, itself a modeling step, that can be said to introduce certain biases. See \citet[683ff]{LuxSch11incl} for a short discussion of the different places bias can enter.}  In particular, the use of \textsc{ERM} relies (and learning-theoretic guarantees for \textsc{ERM} rely) on the problem assumption of stochastic classification that data is sampled i.i.d.\ (this can be extended to a stationarity assumption but not much beyond). For this learning problem, and in particular due to the i.i.d.\ assumption, the ``uniformity meta-bias'' of \textsc{ERM} is provably good, and the ``anti-uniformity meta-bias'' of anti-\textsc{ERM} is provably not. In general, in the same way that any \textsc{NFL} statement concerns a certain learning problem (recall sect.\ \ref{ssec:nounivalgo}),
%PETER to TOM 17-5: added (ik vond het wel belangrijk om 
%expliciet aan te geven dat NFL ook uitgaat van iid, maar maybe it hurts more than it helps, voe je vrij dit weer weg te halen
%and the standard, classic NFL statements in particular rely on the i.i.d. framework and thus assume i.i.d. data
%
any learning guarantee concerns a certain learning problem. Thus our claim is more precisely that many standard learning methods, also \emph{relative to} the learning problem they were designed for, have a model-relative justification.

%TOM to PETER: ben je het eens met het bovenstaande?
%PETER2 ja - met kanttekening dat het een beetje vaag blijft 
% wat een framework dan precies is, maar laten we het niet
% nog verder uitdiepen, we geven al erg veel nuance overal

Finally, recall from our discussion in sect.\ \ref{ssec:local} that it is far from clear that a local conception of induction brings us closer to an absolute, \emph{global} justification of inductive inferences. Similarly, a model-relative justification still leaves the justification for the model in any particular application of a learning algorithm, and indeed the further assumptions encoded in the very formulation of the learning problem. A global justification for the conclusions of a machine learning algorithm must also include the justification for all these assumptions. The obvious threat is an endless justificatory regress, where the motivation for these local assumptions leads us to an earlier inference that itself relies on inductive assumptions that want justification. Note, though, that this regress will soon, if not immediately, lead us to assumptions that we have not actually arrived at by machine learning methods. We will soon have left the domain of machine learning, and face the problem of induction in its full generality. Rather, therefore, than understanding the  \textsc{NFL} theorems as somehow deepening Humean skepticism, the more sober conclusion is that the question of the global justification for the conclusions of machine learning algorithms reduces to the original problem of induction.

\section{Conclusion}\label{sec:concl}
The \textsc{NFL} theorems are commonly understood to show that every learning algorithm must possess a certain bias, and must ultimately lack justification because any such bias must. We have argued that for many standard learning algorithms, this is turning things on their head. \textsc{NFL} results do show that any \emph{data-only} algorithm must have an inherent bias. Presented such an algorithm, we could expose its bias, and question the justification for this bias and thereby for the algorithm. However, many standard learning algorithms are better conceived of as \emph{model-dependent}. The need for a choice of bias is accommodated by such an algorithm from the start: on each application, one must equip it with a particular model, that represents the bias. The algorithm \emph{itself} is generic in that it  does not itself come with a bias: on each application, one must provide it with one. What is more, such algorithms can have a \emph{model-relative} justification: relative to any given model, such an algorithm performs well. Learning-theoretic guarantees show that in that sense some standard learning algorithms \emph{are} sensible, \emph{are} justified---and other possible algorithms are not. This is perfectly consistent with the valid lesson of \textsc{NFL} results that any data-only learning method, including a model-dependent algorithm \emph{plus} a particular choice of model, must possess a bias.

In the course of our argument, we drew some parallels to the broader philosophy of induction. Most importantly, we discussed the role of a general postulate on the induction-friendliness of the world, and the local view of induction that challenges the cogency of such a postulate. We think of our emphasis on the model-dependence of many standard learning algorithms as an instance of the local view of induction. It is important to note, however, that the local view does not yet suffice to escape Hume's skeptical argument, and neither does the model-relative conception in the context of machine learning algorithms. Namely, an \emph{absolute} justification for the conclusions of inductive inferences still requires a justification for the preceding choice of local assumptions or model. 

For that reason, the local view of induction also does not suffice to fully explain the success of our inductive inferences. %, of how we can ``reconcil[e] the continuing failure to rebut Hume's argument with the undoubted fact that induction not only seemed to work but to work surpassingly well'' \citep[10]{How00}. 
Analogously, Wolpert \citeyearpar[1364]{Wol96nc} points at ``a  rather  profound (if somewhat philosophical) paradox,'' that is not yet resolved by the model-dependent perspective on learning algorithms: ``How is it that we perform inference so well in practice, given  the NFL theorems and the limited scope of our prior knowledge?'' That this is not merely a ``somewhat philosophical'' issue is demonstrated, for instance, by the recent debate surrounding the ``paradox of deep learning'' \citep{ZBHRV17iclr,NBMS17nips,AJBKBKMFCBLJ17icml,KawKaeBenxxinc}, which revolves around the perceived lack of a good explanation for the empirical success of deep neural networks. The case of deep learning is particularly interesting, as a clean separation of method and model is here much more contentious, %(see appendix \ref{ssec:appnuannotmodel}),
 and the remaining question of justification does not clearly center on the motivation for a well-articulated choice of model. %If only for this reason, this presents an intriguing case study for the philosophy of induction, too.
%TOM to PETER: ik sluit dus af met het voor ons lastige geval van DNN's. ik denk dat het goed is om aan het einde nog even de verbinding met de filosofie van inductie te maken, maar mijn verborgen agenda hier is ook dat ik overweeg om DNNs onderwerp van mijn veni te maken. zeg het vooral als je deze wijze van conclusie hier niet zo ziet zitten!
%PETER2 Ik vind het goed. Het is een beetje risky omdat we het standbeeld dat we hebben opgezet toch weer een klein beetje omver trekken, maar tegelijkertijd maakt het duidelijk dat we niet alleen nfl-bashing aan het doen zijn maar echt constructief over deze dingen nadenken. Dus : positief!

\vspace*{5mm}

\small

\appendix

\section{The original NFL results and learning theory}\label{app:orignfl}
%PETER2 na lang nadenken heb ik besloten deze hele subsectie op te hangen aan de 'paradox' die jij, en Wolpert, en Teemu Roos, en ik, allemaal gezien hadden. Wat jij suggereerde ging al die kant op maar ik heb het nog wat centraler gemaakt. Dat levert het vlotste verhaal op, denk ik. Wel erg technisch maar ik denk dat dit punt nu voor het eerst in de wereldgeschiedenis helder gemaakt wordt :-) 
The central question in this paper concerned the consistency of the negative \textsc{NFL} results with positive results from learning theory. %We argued that the key to resolving this apparent conflict is the difference between data-only and model-dependent alorithms. 
In sect.\ \ref{sssec:modelnfl}, we explained in formal detail how in the specific case of $\textsc{ERM}$, an \textsc{NFL} statement is consistent with a learning-theoretic guarantee for this algorithm. However, we there discussed the \textsc{NFL} variant due to \cite{ShaBen14}, a version of statement \eqref{eq:NFL}, and not the \emph{original} Wolpert-Schaffer result. %As already mentioned there, 
The original case still leaves something to explain.

Consider a finite hypothesis class $\mathcal{F}$, and two different learning algorithms: \textsc{ERM}, turned into a data-only algorithm by  equipping it with hypothesis class $\mathcal{F}$ (write $A_\textsc{ERM}(\mathcal{F})$) and \emph{anti}-\textsc{ERM} with the same  hypothesis class (selecting, for any training sample $S$, an $f\in \mathcal{F}$ with the \emph{worst} empirical error on $S$; write $A_\textsc{a-ERM}(\mathcal{F})$). The Wolpert-Schaffer results tell us that, under a uniform prior over learning situations, for any sample size $n$, both $A_\textsc{ERM}(\mathcal{F})$ and $A_\textsc{a-ERM}(\mathcal{F})$ have the same expected \textsc{ots} risk of 1/2. Yet the learning theoretic guarantee mentioned in the main text and proved below says that the expected \textsc{iid} risk of $A_\textsc{ERM}(\mathcal{F})$ is not more than $\min_{f^* \in \mathcal{F}}L_\mathcal{D}(f^*) +  \sqrt{|\log \mathcal{F}|/(2n)}$, whereas the expected \textsc{iid} risk of $A_\textsc{a-ERM}(\mathcal{F})$ is not less than $\max_{f_* \in \mathcal{F}}L_\mathcal{D}(f_*) - \sqrt{|\log \mathcal{F}|/(2n)}$. 
%This holds under {\em all\/} distributions (learning situations), so it must also hold under a uniform prior over them. 

This presents us with something of a paradox: it appears that $A_\textsc{ERM}(\mathcal{F})$ and $A_\textsc{a-ERM}(\mathcal{F})$ behave equally well under the uniform prior on learning situations (NFL result), whereas one behaves much better than the other (learning theory) under arbitrary priors, including the uniform. 
Similar paradoxes have been noted before, and it has been suggested that the reason the former negative result and the latter positive result can exist together is that the latter and not the former relies on \textsc{iid} error \citep[1368f]{Wol96inc}. But this contradicts  \cite{RGMT05nips}, who show  that in many realistic learning situations the \textsc{iid} and \textsc{ots} error are very close to each other. 

Here, we show how to resolve the paradox and reconcile both results in situations in which both types of errors are essentially equally large. In \ref{ssec:appwolots} we discuss \textsc{ots} and \textsc{iid} risk and in \ref{ssec:appwolcons} we explain how the paradox is resolved. In \ref{ssec:derivation} we provide a short proof of the relevant learning-theoretic bounds.
% --- the relevance of this being that it will reinforce and clarify some of our findings.
%PETER2 formulering van mij hierboven over relevance is misschien niet heel mooi, misschien kun je er wat beters van maken? 
%TOM2: ik heb in het begin van de sectie een korte motivatie gegeven.

\subsection{OTS and IID risk}\label{ssec:appwolots}
%TOM to PETER: ik stel me voor dat we in deze sectie wat zeggen over de relatie tussen ots en idd, met name jullie NIPS paper, en misschien ook kort over de motivatie voor de ene of de andere. 
%[ik zeg daar al wat over in sectie 2.2 (ots is schijnbaar natuurlijker in het deterministische geval, maar in het stochastische geval is er voor ongeziene instanties ook nog steeds een schattingsprobleem, wat pleit voor iid); ik weet niet of je het daar mee eens bent, maar dat kan misschien ook hier] 
%doel van deze uitleg is (vooral) een basis voor de volgende sectie over de formele consistentie van nfl met leertheorie, misschien dus ook wel beter om alles in één sectie te doen
%PETER2 dit is inderdaad wat ik hier nu doe
First recall that both the NFL and the learning theory setting assume that data $(X_1,Y_1), (X_2,Y_2), \ldots, (X_n,Y_n)$ are i.i.d.\ $\sim \mathcal{D}$ where $\mathcal{D}$ is some distribution on instances in $\mathcal{X}$ and labels in $\mathcal{Y}=\{0,1\}$. For given $\mathcal{D}$, let $\mathcal{D}_X$ denote the marginal distribution on $\mathcal{X}$ and $\mathcal{D}_{Y|X}$ the conditional distribution for $Y$ given $X$ (that is, for each $x \in \mathcal{X}$, 
$\mathcal{D}(Y \mid X=x)$ is a distribution on $\mathcal{Y}$). With this notation we have that $X_1, X_2, \ldots, X_n$ are i.i.d. $\sim \mathcal{D}_X$ (we simply write $X^n \sim \mathcal{D}_X$, leaving independence implicit and abbreviating $(X_1, \ldots, X_n)$ as $X^n$). Similarly, given $X_1, \ldots, X_n$, we have that $Y_1, \ldots, Y_n$ are independent with $Y_i \sim \mathcal{D}_{Y|X} \mid X_i$ (we simply write $Y^n \sim \mathcal{D}_{Y|X} \mid X^n$). 

Based on an analysis of $\mathcal{D}_X$, \cite{RGMT05nips} show that in practically realistic settings, \textsc{iid} and \textsc{ots} risk are often (though not always) extremely close, and analyses pertaining to the one transfer to the other.  To get a first idea of why this might be so, assume that the instance space $\mathcal{X}$ is continuous and $\mathcal{D}_X$ has a probability density on $\mathcal{X}$. Then, for $i \neq j$,  $\mathcal{D}_X(X_i= X_j) =0$: the probability that we see the same feature vector twice is $0$, and more generally, for any finite sample $S$, the probability that an independently sampled test instance $X \sim D_X$ is contained in $S$ is also $0$. Since both \textsc{iid} and \textsc{ots} risk involve an expectation over such an independent $X$, this implies that for continuous $\mathcal{X}$, we must have  that  \textsc{iid} and \textsc{ots} risk coincide:  $L_{\mathcal{D} \setminus S}(A(S)) = L_{\mathcal{D}}(A(S))$, almost-surely under $\mathcal{D}$. 

Now the original \textsc{NFL} results require $\mathcal{X}$ to be finite or countable, so the above reasoning does not necessarily hold. But if $\mathcal{X}$ is finite but not too small, and equipped with a uniform distribution $\mathcal{D}_X$, then the situation is quite similar to the continuous setting (the probability that an $X$ contained in $S$ is also contained in a test set is \emph{almost} zero) and we can still show that, at sample sizes of interest, the difference between both quantities is negligible. 
For example, suppose that features are $m$-dimensional binary vectors, $\mathcal{X} = \{0,1\}^m$, $\mathcal{D}_X= \mathcal{U}_X$ is uniform, and the sample size is $n$. Then lemma~1 of \cite{RGMT05nips} gives that, under every distribution $\mathcal{D}_{Y|X}$,
\begin{equation}
\left|\Expec_{X^n \sim \mathcal{U}_X, Y^n \sim \mathcal{D}_{Y|X} \mid X^n }
\left[ L_{\mathcal{D} \setminus S}(A(S))-  L_{\mathcal{D}}(A(S) )\right] \right| \leq n 2^{-m}. \label{eq:iidisots}
\end{equation}

So, for example, if $m \geq 40$ (40 covariates being much less than what is common in modern machine learning) 
and sample size $n$ is less than $10^6$ (as is the case in many realistic scenarios), then the difference between the expected behaviour, over training samples, of the two risk measures is less than approximately $10^{-6}$, which is completely negligible for practical purposes. It is true that in practice, $\mathcal{D}_X$ will usually not be uniform. But \cite{RGMT05nips} show that even for highly nonuniform $\mathcal{D}_X$, both error measures are often very close---the closeness can even be estimated from the obtained sample $S$.

So in those (realistic) situations where $\textsc{iid}$ and $\textsc{ots}$ essentially coincide, how do we account for the co-existence of the \textsc{NFL} results and positive learning guarantees? %We now proceed to resolve this apparent paradox.

\subsection{Consistency of the Wolpert-Schaffer results with learning theory}\label{ssec:appwolcons}
%TOM to PETER: hier stel ik me voor dat we kort laten zien hoe:
%-1- de originele nfl (mét uniformiteitsaanname) consistent is met de pacgarantie in de hoofdtekst, en dat dit niet zoveel met ots v iid te maken heeft:
%dat is dus de truc uit je email om nfl's in de limiet naar een continue domein (waar ots=idd) te bekijken, en op te merken dat dit alleen kan als elke hypothese het niet beter doet dan random gokken. dus herhaling van het punt van sectie 2.4: uniformiteitsaanname is echt een "onleerbaarheids"aanname
%-2- de "verfijnde" nfl van sectie 2.5 (dus zónder uniformiteitsaanname) consistent is met de pacgarantie: hier is de observatie dat er voor gegeven ERM+F en a-ERM+F distributies D moeten bestaan waaronder de laatste beter is dan de eerste, maar dat moeten D zijn waarvoor de klasse F heel oninteressant is: alle hypotheses zijn ongeveer even goed
%In order to investigate the relation between the original \textsc{NFL} and learning theory results further, we must 
We first restate the \textsc{NFL} result in a fully precise manner. Assume that $\mathcal{X}$ is finite or countable, and recall that $S=((X_1,Y_1), \ldots, (X_n,Y_n))$. Then for every distribution $\mathcal{D}_X$ on $\mathcal{X}$, 
\begin{equation}\label{eq:nflprecise}
\Expec_{\mathcal{D}_{Y|X} \sim \mathcal{U}} \Expec_{ X^n \sim \mathcal{D}_X, Y^n \sim \mathcal{D}_{Y|X} \mid X^n }
\left[ L_{\mathcal{D} \setminus S}(A(S))\right]=1/2,
\end{equation} where $\mathcal{U}$ is the uniform distribution on conditional distributions $\mathcal{D}_{Y|X}$. To be clear, this is the distribution such that conditional on each $x \in \mathcal{X}$, $p_{|x} := \mathcal{D}(Y=1 \mid X=x)$ has a uniform distribution on the unit interval $[0,1]$.
%PETER ADDED 14-5 to respond to remark 1(a) of reviewer 1
To see that this is a natural definition of ``uniform'' in this context, note that $p_{|x}$ indicates the mean of $Y$ given $x$ according to $\mathcal{D}$, and $\mathcal{U}$ is thus also uniform on the mean. 

In contrast to (\ref{eq:nflprecise}), we derive below that for every distribution $\mathcal{D}_X$, and for every distribution $\mathcal{U}'$ on $\mathcal{D}_{Y|X}$ (including the uniform \textsc{NFL} distribution $\mathcal{U}$ defined above), 
%noting that $S=((X_1,Y_1), \ldots, (X_n,Y_n))$,
\begin{align}
& \Expec_{\mathcal{D}_{Y|X} \sim \mathcal{U}'} \Expec_{ X^n \sim \mathcal{D}_X, Y^n  \sim \mathcal{D}_{Y|X} \mid X^n }
\left[ L_{\mathcal{D}}(A_{\text{\sc ERM}}(\mathcal{F}, S ) ) - \min_{f^* \in \mathcal{F}} L_{\mathcal{D}}(f^*) \right] \leq \sqrt{\frac{\log |\mathcal{F}|}{2 n}} \label{eq:mini} \\
& \Expec_{\mathcal{D}_{Y|X} \sim \mathcal{U}'} \Expec_{ X^n \sim \mathcal{D}_X, Y^n \sim \mathcal{D}_{Y|X} \mid X^n }
\left[ \max_{f^\circ \in \mathcal{F}} L_{\mathcal{D}}(f^\circ) -  L_{\mathcal{D}}(A_{\text{\sc a-ERM}}(\mathcal{F},S) \right] 
\leq \sqrt{\frac{\log |\mathcal{F}|}{2 n}}. \label{eq:maxi}
\end{align}

Since both the NFL and learning theory results hold for every distribution on finite or countable $\mathcal{X}$, we get an instance of our paradox if we (a) take a distribution $\mathcal{D}_X$ for which $\textsc{iid}$ and $\textsc{ots}$ error differ by a negligible amount, say $\delta$ very close to $0$, at the given sample size $n$, and (b) a combination of $\mathcal{F}$ and $n$ for which $\sqrt{(\log |\mathcal{F}|)/2n}$ is very close, say $\epsilon$, to $0$, so that the bounds (\ref{eq:mini}) and (\ref{eq:maxi}) are non-void. We henceforth call any combination $(\mathcal{D}_X,|\mathcal{F}|,n)$  for which both (a) and (b) are the case an $(\epsilon,\delta)$- {\em paradoxical learning situation}, with the understanding that the closer $\epsilon$ and $\delta$ to $0$, the more paradoxical. 

To be in an $(\epsilon,\delta)$-paradoxical situation, we see that it is sufficient, for any finite $\mathcal{F}$, to take $n$ sufficiently large, and, from~(\ref{eq:iidisots}), to take $\mathcal{X}$ finite and  $\mathcal{D}_X$ uniform, with $m$ and $n$ so that $\delta = n 2^{-m}$ is negligibly small. We thus already know that such situations exist, and the result of \cite{RGMT05nips} implies that they exist for much more general $\mathcal{D}_X$ as well.  Note that only the size of $\mathcal{F}$, not the definitions of its constituent $f$'s, is relevant to determine whether the situation is paradoxical. 

How can we reconcile (\ref{eq:nflprecise}), (\ref{eq:mini}) and (\ref{eq:maxi}) in paradoxical learning situations? 
%PETER3 If the right-hand sides of (\ref{eq:mini}) and 
%(\ref{eq:maxi}) were $0$, there would be a real 
% contradiction. The fact that they are not $0$ resolves the paradox: 
The key observation is that there would only be a real contradiction if the 
optimal classifier $f^* \in \mathcal{F}$ in our class and the worst classifier $f^\circ \in \mathcal{F}$ in our class
differed substantially in terms of their risk $L_{\mathcal{D}}$. 
Thus, rather than being contradictory, taken together,  
(\ref{eq:nflprecise}), (\ref{eq:mini}) and (\ref{eq:maxi}) simply imply that, in $(\epsilon,\delta)$-paradoxical situations,  under the \textsc{NFL} prior, {\em no matter what $\mathcal{F}$ of given size we chose}, we expect both the optimal classifier $f^* \in \mathcal{F}$  and the worst classifier $f^\circ \in \mathcal{F}$ to have both $\textsc{iid}$ and $\textsc{ots}$ error within about 
$\epsilon (1+ \delta)$ of $1/2$. Thus, in the paradoxical cases in which $\epsilon,\delta$ are very small,  both $f^*$ and $f^{\circ}$ behave essentially no better or worse than random guessing. As a consequence, in paradoxical situations, under the NFL prior, we expect $\mathcal{F}$, no matter how we chose it, to be essentially useless: it contains no useful (classification error substantially smaller than $1/2$) $f$, hence ``there is nothing to be learned from  $\mathcal{F}$.'' This reinforces the point made in sect.\ \ref{ssec:arequiv} that the \textsc{NFL} prior is a prior under which learning is impossible: at least in paradoxical situations, it makes it next to impossible that any $\mathcal{F}$ we might choose provides us with anything to learn.

Next we move to a variation of the paradox:  consider the instance of \textsc{NFL} statement \eqref{eq:NFL} that is implied by the Wolpert-Schaffer result, saying that for any two data-only algorithms, there is a learning situation $\mathcal{D}$ such that the first algorithm has expected \textsc{ots} risk at least 1/2, while the second has expected \textsc{ots} risk strictly below 1/2.
Again, this may appear paradoxical if we apply this result to 
$A_\textsc{ERM}(\mathcal{F})$ and $A_\textsc{a-ERM}(\mathcal{F})$,
but now the paradox may seem more serious because it does not refer to the \textsc{NFL} prior.  But again, the statement can be reconciled with the learning-theoretic guarantees (\ref{eq:mini}) and (\ref{eq:maxi}).
%PETER3
% by noting that the right-hand side of these equations is %not $0$ but rather a small number. 
Rather than contradicting (\ref{eq:NFL}), in conjunction with (\ref{eq:NFL}) and (\ref{eq:iidisots}), they imply that 
\begin{enumerate}\item
there do exist $(\epsilon,\delta)$-paradoxical learning situations with very small $\epsilon$ and $\delta$ in which 
both the  \textsc{iid} and \textsc{ots} error of $A_\textsc{a-ERM}(\mathcal{F})$ is smaller than $1/2$ while both 
\textsc{iid} and \textsc{ots} error of $A_\textsc{ERM}(\mathcal{F})$ are larger than $1/2$ (``anti-\textsc{ERM} is better than random guessing, \textsc{ERM} is worse''), $\ldots$
\item $\ldots$ yet in such situations $\mathcal{F}$ must be essentially useless in the sense that both \textsc{iid} and \textsc{ots} errors of its best element $f^*$  and its worst element $f^{\circ}$ are within about  $\epsilon$ of $1/2$: the {\em only\/} $(\epsilon,\delta)$-paradoxical learning situations in which anti-\textsc{ERM} is better than random guessing and outperforms \textsc{ERM} must concern hypothesis classes $\mathcal{F}$ that only contain $f$ that are themselves at most $\epsilon$-better than random guessing (and then anti-\textsc{ERM} is itself also at most $\epsilon$ better than random guessing). 
\end{enumerate}
Viewed in this way, the paradoxical situations become much less paradoxical.

\subsection{Derivation of expected risk bounds}\label{ssec:derivation}
We now give a compact  derivation of the following result:
{\em for all distributions $\mathcal{D}_X$ on $\mathcal{X}$, all conditional distributions $\mathcal{D}_{Y \mid X}$ on $\mathcal{Y}$ given $X$, with $\mathcal{D}$ denoting the corresponding joint distribution on $\mathcal{X} \times \mathcal{Y}$ and $S= ((X_1,Y_1), \ldots, (X_n,Y_n))$, we have:} 
%PETER2 we kunnen hier evt een Proposition 1 van maken maar ik weet niet of dat opportuun is in een filosofisch tijdschrift...
\begin{align}
&\Expec_{ S \sim \mathcal{D}}
\left[ L_{\mathcal{D}}(A_{\text{\sc ERM}}(\mathcal{F},S)) - \min_{f^* \in \mathcal{F}} L_{\mathcal{D}}(f^*) \right] = 
\nonumber \\
& \label{eq:minib} 
\Expec_{ X^n \sim \mathcal{D}_X, Y^n  \sim \mathcal{D}_{Y|X} \mid X^n }
\left[ L_{\mathcal{D}}(A_{\text{\sc ERM}}(\mathcal{F}, S) ) - \min_{f^* \in \mathcal{F}} L_{\mathcal{D}}(f^*) \right] \leq \sqrt{\frac{\log |\mathcal{F}|}{2 n}}.
\end{align}

Since (\ref{eq:minib}) holds for all conditional distributions $\mathcal{D}_{Y \mid X}$, it must also hold in expectation over every prior distribution $\mathcal{U}'$ on $\mathcal{D}_{Y \mid X}$, so that (\ref{eq:mini}) follows. Then \eqref{eq:maxi} follows by repeating all the steps of the proof below with obvious modifications. 
The result for two-fold forward-validation in the main text follows from (\ref{eq:minib}) as follows. First, let $\mathcal{F} = \{\hat{f}_1, \ldots, \hat{f}_{m} \}$ be the $m$ classifiers that were output by the $m$ algorithms $A_1, \ldots, A_{m}$ on $S_1$, the first half of the sample. Now, use the result above with $S_2$ in the role of $S$, conditional on $S_1$. Then $\mathcal{F}$ is fixed, and $n$ gets replaced by $n/2$, and the result follows by further using that the expectation (over $S_1$) of a minimum is no larger than the minimum of the expectation.

{\em Proof.\/}
Denote for each fixed classifier $f \in \mathcal{F}$, the loss it makes on the $i$-th outcome by $\ell_{f}(X_i,Y_i)$ and one minus this loss as $\ell'_{f}(X_i,Y_i)$. 
Then $\ell'_{f}(X_1,Y_1), \ell'_{f}(X_2,Y_2), \ldots$ is i.i.d.\ Bernoulli. Let  $(X,Y)$ be  another i.i.d.\ copy of $X_1,Y_1$ (think of it as a ``test'' example).    
By Hoeffding's inequality \citep[56]{ShaBen14} we have, for all $\eta > 0$, that  ${\bf E}\left[ 
\exp(\eta\sum_{i=1}^n (\ell'_f(X_i,Y_i) - {\bf E}[\ell'_f(X,Y)] ) 
)\right]\leq \exp(\eta^2 n/8)$ so that also, for any learning algorithm $A$ that outputs, upon seeing sample $S= (X_1,Y_1), \ldots, (X_n,Y_n)$, a classifier $A(S) \in \mathcal{F}$, we have 
\begin{align*}
& {\bf E}\left[ 
\exp(\eta \sum_{i=1}^n (\ell'_{A(S)}(X_i,Y_i) - {\bf E}[\ell'_{A(S)}(X,Y)] ) 
)\right] \\
\leq \ &  
{\bf E}\left[ \exp(\max_{f \in \mathcal{F}}  \eta \cdot \sum_{i=1}^n (\ell'_f(X_i,Y_i) - {\bf E}[\ell'_f(X,Y)] ) 
)\right] \\
= \ & \max_{f \in \mathcal{F}} 
{\bf E}\left[ \exp(\eta \sum_{i=1}^n (\ell'_f(X_i,Y_i) - {\bf E}[\ell'_f(X,Y)] ) 
)\right] \\
\leq \ & \sum_{f \in \mathcal{F}} 
{\bf E}\left[ \exp(\eta \sum_{i=1}^n (\ell'_f(X_i,Y_i) - {\bf E}[\ell'_f(X,Y)] ) 
)\right] \leq 
 \exp( \log |\mathcal{F}| +\eta^2 n/8).
\end{align*}
Jensen's inequality, division by $\eta n $, using that the result holds for all $\eta>0$, and differentiation, now give that: 
$$
{\bf E}\left[\frac{1}{n} 
\sum_{i=1}^n (\ell'_{A(S)}(X_i,Y_i) - {\bf E}[\ell'_{A(S)}(X,Y)] ) 
\right] 
\leq \min_{\eta > 0} \ \left\{ \frac{\eta}{8} + \frac{\log | \mathcal{F} |}{\eta n} \right\} = \sqrt{\frac{\log | \mathcal{F}|}{2 n}}.
$$
If we take $A(S) = A_\textsc{ERM}(\mathcal{F},S)$ to be an instance of \textsc{ERM} applied to $\mathcal{F}$, and replace $\ell = 1- \ell'$, this gives
\begin{align*}
& {\bf E}_{S \sim \mathcal{D}}[ {\bf E}_{(X,Y) \sim \mathcal{D}} [\ell_{A_\textsc{ERM}(\mathcal{F},S)}(X,Y)] 
- {\bf E}_{(X,Y) \sim \mathcal{D}} [\ell_{f^*}(X,Y)] ] 
\leq \\
& 
{\bf E}_{S \sim \mathcal{D}} \left[\frac{1}{n} 
\sum_{i=1}^n \ell_{A_\textsc{ERM}(\mathcal{F},S)}(X_i,Y_i)  \right]
- {\bf E}_{(X,Y) \sim \mathcal{D}} [\ell_{f^*}(X,Y)] + \sqrt{\frac{\log | \mathcal{F}|}{2 n}} = \\
& 
{\bf E}_{S \sim \mathcal{D}} \left[\frac{1}{n} 
\sum_{i=1}^n \ell_{A_\textsc{ERM}(\mathcal{F},S)}(X_i,Y_i)  
- \frac{1}{n} \sum_{i=1}^n \ell_{f^*}(X_i,Y_i)] + \sqrt{\frac{\log | \mathcal{F}|}{2 n}} \right] \leq \sqrt{\frac{\log | \mathcal{F}|}{2 n}},
\end{align*}
where we used that $\ell_{f*}(X_i,Y_i)$ is an i.i.d.\ random variable, and the fact that \textsc{ERM}'s risk on the training data can by definition not be larger than that of $f^*$. This
shows (\ref{eq:minib}). \QEDP

%\subsubsection{Off-Training Set vs. Classification Error} PETER: STILL TO ELOBARE THIS. CURRENT PLAN IS SOMETHING LIKE: first explain difference 'classification error' ($=$ Wolpert's iid error) and off-training set error. Explain that tension between 'ERM good/anti ERM bad' and NFL results can be resolved by noting that with the use of off-training set error and a uniform prior, the future is completely 'decoupled' from the past. In contrast, we follow the standard approach of learning theory, in which the future is coupled to the past via a fixed (but arbitrary!!) underlying distribution. This means we have to use the standard i.i.d. error (Note that the Shalev-Shwartz-Ben-David NFL example also uses this iid error). But this is completely standard and everybody is happy with it, so we do not think it needs further justification; note also that if $\mathcal{X}$ is a continuum with a probability density on it, the two types of error measures are the same anyway; and if $\mathcal{X}$ is a very large space (e.g. a continuum discretized to a fixed nr of bytes, as very often happens in practice), then off-training and classification error are still closely tied together  (see Roos et al. for further discussion).  

\section{Model-dependent algorithms: Some details and nuances}\label{app:details}
Here we discuss some limitations of standard model-relative learning algorithms (\ref{ssec:appnuanlims}), and the existence of learning algorithms that are not clearly model-dependent (\ref{ssec:appnuannotmodel}).

\subsection{Even good learning algorithms have limitations}\label{ssec:appnuanlims} 
Our point in sect.\ \ref{ssec:models} is emphatically not that (pragmatic) Bayes, \textsc{ERM} or cross-validation are perfect methods. %For completeness, we list here a few shortcomings.

In the case of \textsc{ERM}, we already mentioned that it is not suitable for $\mathcal{F}$ that are very complex or large relative to the given training set size $n$. 
More generally, in this regime it can sometimes behave in surprisingly bad ways, with \textsc{ERM}'s risk temporarily increasing in the small $n$-regime \citep{loog2019minimizers}. Additionally, even if the goal is to learn a classifier with small classification $(0/1)$-error, \textsc{ERM} often delivers better results if applied on the training set with a different, {\em proxy\/} loss function that has nicer mathematical properties. This includes logistic and hinge loss, which avoid the discontinuites of the $0/1$-loss.  
 
As to cross-validation, it can perform poorly when the bound stated above does not hold, i.e., if the number of constituent algorithms grows exponentially in $n$. This happens, for example, in variable selection problems. For these, {\em penalized\/} \textsc{ERM} approaches such as the Lasso are more suitable \citep{HasTibFri09}. 
%PETER to TOM: below are some shameless self-references but it might help to make clear that I am actually an EXPERT on writing about limitations and bad behaviour of learning algorithms and STILL I disagree with Wolpert...
%TOM to PETER: this is great! (We will need to rephrase this slightly before submitting for blind review, but we can change it back afterwards for the final version)
Moreover, if we adopt different loss functions, such as the squared error loss, typical in regression, or the logarithmic loss in density estimation, the picture changes completely. Whether or not cross-validation is the right approach can then depend to some extent on the goal of the inference \citep{Yang05a,erven2012catching}: does one want to learn the best squared-error predictor (cross-validation, behaving like \textsc{AIC}, is often suitable), or does one want to find out which components of the vector $X$ are correlated with $Y$ (cross-validation is not suitable)? 

There are also inherent differences in the type of assumption codified into a probabilistic model $\mathcal{M}$, as in Bayesian machine learning, or a class $\mathcal{F}$ as in \textsc{ERM}, which further point to necessary conditions for the algorithms to work well. For example, if the employed probabilistic model is wrong-but-useful (that is to say, if it contains a distribution that leads to good predictions for the prediction task of interest), the inductive assumption underlying pragmatic Bayesian inference does not hold and in some cases Bayesian inference does fail dramatically in practice, both in regression and classification \citep{GrunwaldO17,GrunwaldL07}. Specifically, in such cases it can happen that no matter how many data are observed, the Bayesian posterior never concentrates around the best-predicting distribution in the model. In contrast, the inductive assumption underlying \textsc{ERM} is of a much more agnostic type, in which hypotheses are not assumed to be ``true'' but just ``useful'' (have small classification error), so by construction there can be no problems with ``wrong-but-useful'' models.

Indeed (viz.\ the references above), one of us has published extensively on the limitations and suitable domain of application of methods such as pragmatic Bayes, \textsc{ERM} and cross-validation. But acknowledging that these methods have limitations is very different from saying they lack any inherent justification. In fact, studying their strengths and limitations is done with, and shows the usefulness of, machine learning theory. An \textsc{NFL} claim that ``all algorithms are equally (un)justified in principle'' seems to preclude such study.

\subsection{Algorithms that are not clearly model-dependent}\label{ssec:appnuannotmodel}
Some popular learning algorithms that appear to be data-only can be understood as being model-dependent with a particular model already filled in. A prototypical example are support vector machines (\textsc{SVM}'s, \citealp{Vap98})  with a particular choice of \emph{kernel}. While commonly viewed as ``machines'' that take merely data as input, they can be recast as a model-dependent algorithm: the model is a hypothesis class $\mathcal{F}$ of linear combinations of basis functions $\phi_1(X), \phi_2(X),\ldots$ of the instances, with the $\phi_j$ implicitly given by the kernel. The learning algorithm is penalized \textsc{ERM} with the hinge loss, a ``proxy'' for the $0/1$-loss: it picks the $f \in \mathcal{F}$ that minimizes the hinge loss on the training sample with a penalty for the $L_1$-norm of the parameters \citep{HasTibFri09}.

For other data-only learning algorithms, however, such a decomposition is more tenuous. For example, it is not immediately clear whether one can recast the nearest-neighbor algorithm as such. (Though see \citealp[666]{LuxSch11incl} for a discussion of $k$-nearest neighbor as working with a certain function class determined by the parameter $k$.)
We stress that this does not contradict our main point: we merely state that {\em many\/} (that is to say, {\em not all!})\ often-used learning methods are inherently model-dependent, and for those, claims that they work well should be understood relative to the given models. 
%TOM to PETER: in the 2011 overview by Von Luxburg and Schölkopf they do analyse n-nearest neighbor in terms of hypothesis classes for ERM---maybe mention?
%PETER2 ja
%TOM to PETER:moreover, there does of course exist a "universal consistency" guarantee for k-NN, so even here a learning theoretic justification is possible. perhaps the best way to see it is that biases already enter in the problem set-up (choice of metric), and the guarantee is relative to those. mention or is this all taking us too far?
%PETER2 deze tweede opmerking brengt ons inderdaad te ver, denk ik. Dan zouden we over metrics-as-models moeten praten en zo...ik zou het maar gewoon zo laten. 

Finally, for some instances of \textsc{ERM}, though clearly a model-dependent procedure, an additional complication arises. What we have in mind here is deep learning, where $\mathcal{F}$ is the set of all neural networks with a given structure, parameterized by their weights. The standard learning algorithm in this setting is stochastic gradient descent (\textsc{SGD}), which is usually  iterated until the error on the training set is zero.  This makes \textsc{SGD} an instance of \textsc{ERM}, but it is a very special one. Neural nets typically allow for a multitude (billions) of different local minima in weight space, all with empirical error zero, but \textsc{SGD} directs learning towards very particular minima. For example, these minima tend to be ``broad'' (small perturbations of weights do not cause a noticeable change in predictions) or equivalently, the weights can be grossly discretized and hence the description of the network shortened without sacrificing accuracy \citep{dziugaite2017computing}. Moreover, the found weight vector tends to be  small under a particular, nonstandard norm on vector spaces \citep{bartlett2017spectrally,neyshabur2015norm}. 

Thus, in the use of \textsc{ERM}-by-\textsc{SGD} with a neural network model there is a complicated interaction between the learning algorithm and the model: the same model trained with different instances of ERM that end up in very different minima might lead to very different
%TOM to PETER: instead of "bad", "very different"?
 generalization behaviour. Moreover, the models typically have so many weights that they can represent basically any continuous function from input to output, so they are much too large or complex to represent our inductive bias. Rather, it seems that in those instances of learning problems on which deep learning works so well, \textsc{SGD} has a tendency to find a solution in a particularly ``simple'' (small norm, broad minima, compressible) subset of weight space. There is therefore an interplay between the learning algorithm (\textsc{SGD}), the ``true'' inductive bias (namely, those problems in which deep learning works well)
%TOM to PETER: dit lijkt dan wel weer dicht bij een Wolperiaanse ware prior over de structuur van de wereld te komen---we hoeven er hier niets over te zeggen, maar ik ben benieuwd hoe jij hier naar kijkt?
 and ``effective'' model complexity (for such problems, \textsc{SGD} only explores a tiny fraction of weight space making the model much less complex). 
 
 In sum, in deep learning, there is no crisp separation between inductive bias and learning algorithm. We stress again that this does not invalidate our main point: for algorithms like cross-validation, there {\em is\/} a clear separation, and quality assessments should be done in a model-relative way.

\DeclareRobustCommand{\VAN}[3]{#3} % proper Dutch 'van/de' capitalisation

\bibliographystyle{abbrnvatnoaddress}
%\bibliography{all}

\end{document}